\documentclass[utf8]{frontiersSCNS} 

\usepackage{url,lineno,microtype,caption}
\usepackage[onehalfspacing]{setspace}

\def\keyFont{\fontsize{8}{11}\helveticabold }
\def\firstAuthorLast{Begu\v{s}} 
\def\Authors{Ga\v{s}per Begu\v{s}$^{1,2}$}
\def\Address{$^{1}$Department of Linguistics, University of Washington, Seattle, WA, USA\\
$^{2}$Department of Linguistics, University of California, Berkeley, Berkeley, CA, USA}
\def\corrAuthor{Ga\v{s}per Begu\v{s}}

\def\corrEmail{begus@uw.edu}

\usepackage{amsmath}
\usepackage{amsfonts}
\usepackage{amssymb}
\usepackage{mathrsfs}
\usepackage{graphicx}

\usepackage{multirow}
\usepackage{pifont}

\usepackage{tipa}
\usepackage{tikz}

\begin{document}
\onecolumn
\firstpage{1}

\title[Generative Adversarial Phonology]{Generative Adversarial Phonology: Modeling unsupervised phonetic and phonological learning with neural networks} 

\author[\firstAuthorLast ]{\Authors} 
\address{} 
\correspondance{} 

\extraAuth{}

\maketitle

\begin{abstract}

\section{}
Training deep neural networks on well-understood dependencies in speech data can provide new insights into how they learn internal representations. This paper argues that acquisition of speech can be modeled as a dependency between random space and generated speech data in the Generative Adversarial Network architecture and proposes a methodology to uncover the network's internal representations that correspond to phonetic and phonological properties.  The Generative Adversarial architecture is uniquely appropriate for modeling phonetic and phonological learning because the network is trained on unannotated raw acoustic data and learning is unsupervised without any language-specific assumptions or pre-assumed levels of abstraction. A Generative Adversarial Network was trained on an allophonic distribution in English, in which voiceless stops surface as aspirated word-initially before stressed vowels, except if preceded by a sibilant [s]. The network successfully learns the allophonic alternation: the network's generated speech signal contains the conditional distribution of aspiration duration. The paper proposes a technique for establishing the network's internal representations that identifies latent variables that  correspond to, for example, presence of [s] and its spectral properties. By manipulating these variables, we actively control the presence of [s] and its frication amplitude in the generated outputs. This suggests that the network learns to use latent variables as an approximation of phonetic and phonological representations. Crucially, we observe that the dependencies learned in training extend beyond the training interval, which allows for additional exploration of learning representations. The paper also discusses how the network's architecture and innovative outputs resemble and differ from linguistic behavior in language acquisition, speech disorders, and speech errors,  and how well-understood dependencies in speech data can help us interpret how neural networks learn their representations.

\tiny
 \keyFont{ \section{Keywords:}  generative adversarial networks, deep neural network interpretability, language acquisition, speech, voice onset time, allophonic distribution} 
\end{abstract}

\section{Introduction}
\label{intro}

How to model language acquisition is among the central questions in linguistics and cognitive science in general. Acoustic speech signal is the main input for hearing infants acquiring language. By the time acquisition is complete, humans are able to decode and encode information from or to a continuous speech stream and construct a grammar that enables them to do so \citep{saffran96,saffran06,kuhl10}. In addition to syntactic, morphological, and semantic representations, the learner needs to learn phonetic representations and phonological grammar: to analyze and in turn produce speech as a continuous acoustic stream represented by mental units called \emph{phonemes}.  Phonological grammar manipulates these discrete units and derives surface forms from stored lexical representations. The goal of linguistics and more specifically, phonology, is to explain how language-acquiring children construct a phonological grammar, how the grammar derives surface outputs from inputs, and what aspects of the grammar are language-specific in order to tease them apart from those aspects that can be explained by general cognitive processes or historical developments \citep{lacy06a,dk13,moreton08,mp12a,mp12b,begusDiss}. 

Computational models have been invoked for the purpose of modeling language acquisition and phonological grammar ever since the rise of computational methods and computationally informed linguistics (for an overview of the literature, see \citealt{alderete18,dupoux18,jarosz19,pater19}). One of the major shortcomings of the majority of the existing proposals is that learning is modeled with an already assumed level of abstraction \citep{dupoux18}. In other words, most of the proposals model phonological learning as symbol manipulation of discrete units that already operates on the abstract, discrete phonological level. The models thus require strong assumptions that phonetic learning has already taken place, and that phonemes as discrete units have already been inferred from continuous speech data (for an overview of the literature, see \citealt{oudeyer05,oudeyer06,dupoux18}).

This paper proposes that language acquisition can be modeled with Generative Adversarial Networks  \citep{goodfellow14}. More specifically, phonetic and phonological computation is modeled as the mapping from random space to generated data of a Generative Adversarial Network \citep{goodfellow14} trained on raw unannotated acoustic speech data in an unsupervised manner \citep{donahue19}. To the author's knowledge, language acquisition has not been modeled within the GAN framework despite several advantages of this architecture. The characteristic feature of the GAN architecture is an interaction between the Generator network that outputs raw data and the Discriminator that distinguishes real data from Generator's outputs \cite{goodfellow14}.  A major advantage of the GAN architecture is that learning is completely unsupervised, the networks include no langauge-specific elements, and that, as is argued in Section \ref{experiment} below,  phonetic learning is modeled simultaneously  with phonological learning.   The discussion on the relationship between phonetics and phonology is highly complex \citep{kd94,c06,ks06}. Several opposing proposals, however, argue that the two interact at various different stages and are not dissociated from each other \citep{h99,p01,f16,fruehwald17}. A network that models learning of phonetics from raw data and shows signs of phonological learning is likely one step closer to reality than models that operate with symbolic computation and assume phonetic learning has already taken place independently of phonology (and vice versa). 

We argue that the latent variables in the input of the Generator network  can be modeled as approximates to phonetic or potentially phonological representations that the Generator learns to output into a speech signal by attempting to maximize the error rate of a Discriminator network that distinguishes between real data and generated outputs. The Discriminator network thus has a parallel in human speech:  the imitation principle \citep{nguyen15}.  The Discriminator's function is to enforce that the Generator's outputs resemble (but do not replicate) the inputs as closely as possible. The GAN network thus incorporates both the pre-articulatory production elements (the Generator) as well as the imitation principle (the Discriminator) in speech acquisition. While other neural network architectures might be appropriate for modeling phonetic and phonological learning as well, the GAN architecture is unique in that it combines a network that produces innovative data (the Generator) with a network that forces imitation in the Generator. Unlike, for example, autoencoder networks, the Generative Adversarial network lacks a direct connection between the input and output data and generates innovative data rather than data that resembles the input as closely as possible.  

We train a Generative Adversarial Network architecture implemented for audio files in \cite{donahue19} (WaveGAN)  on  raw speech data that contains information for an allophonic distribution: word-initial pre-vocalic aspiration of voiceless stops ([\textipa{"p\textsuperscript{h}It}] $\sim$ [\textipa{"spIt}]).  The data is curated in order to control for non-desired effects, which is why only sequences of the shape \#TV and \#sTV\footnote{T represents voiceless stops /p, t, k/, V represents vowels (see Figure \ref{timitProc}), and \# represents a word boundary.} are fed to the model. This allophonic distribution is  appropriate for testing learnability in a GAN architecture, because the dependency between the presence of [s] and duration of VOT is not strictly local. To be sure, the dependency is local in phonological terms, as [s] and T are two segments and immediate neighbors, but in phonetic terms, a period of closure intervenes between the aspiration and the period (or absence thereof) of frication noise of [s]. It is not immediately clear whether a GAN model is capable of learning such non-local dependencies. To our knowledge, this is the first proposal that tests whether neural networks are able to learn an allophonic distribution based on raw acoustic data.

The hypothesis of the computational experiment presented in Section \ref{experiment} is the following: if  VOT duration is conditioned on the presence of [s] in output data generated from noise by the Generator network, it means that the Generator network has successfully learned a phonetically non-local allophonic distribution. 
Because the allophonic distribution is not strictly local and has to be learned and actively controlled by speakers (i.e.~is not automatic), evidence for this type of learning is considered phonological learning in the broadest sense.  Conditioning the presence of a phonetic feature based on the presence or absence of a phoneme that is not automatic is, in most models, considered part of phonology and is derived with phonological computation. That the tested distribution is non-automatic and has to be actively controlled by the speakers is evident from L1 acquisition: failure to learn the distribution results in longer VOT durations in the sT condition documented in L1 acquisition (see Section \ref{parallelshuman}).

The results suggest that phonetic and phonological learning can be modeled simultaneously, without supervision, directly from what language-acquiring infants are exposed to: raw acoustic data. A GAN model trained on an allophonic distribution is successful in learning to generate acoustic outputs that contain this allophonic distribution  (VOT duration). Additionally, the model  outputs innovative data for which no evidence was available in the training data, allowing a direct comparison between human speech data and the GAN's generated output. As argued in Section \ref{m12255}, some outputs are consistent with human linguistic behavior and suggest that the model recombines individual sounds, resembling phonemic representations and productivity in human language acquisition (Section \ref{discussion}).

This paper also proposes a technique for establishing the Generator's internal representations.  The inability to uncover networks' representations  has been used as an argument against neural network approaches to linguistic data (among others in \citealt{rawski19}). We argue that the internal representation of a network can be, at least partially, uncovered. By regressing annotated dependencies between the Generator's latent space and output data, we identify values in the latent space that correspond to linguistically meaningful features in generated outputs. This paper demonstrates that manipulating the chosen values in the latent space has  phonetic effects in the generated outputs, such as the presence of [s] and the amplitude of its frication. In other words, the GAN learns to use random noise as an approximation of phonetic (and potentially phonological) representations.  This paper proposes that dependencies, learned during training in a latent space that is limited by some interval, extend beyond this interval. This crucial step allows for the discovery of several phonetic properties that the model learns. 

By modeling phonetic and phonological learning with neural networks without any language-specific assumptions, the paper also addresses a broader question of how many language-specific elements are needed in models of grammar and language acquisition. Most of the existing models require at least some language-specific devices, such as rules in rule-based approaches or pre-determined constraints with features and feature matrices in connectionist approaches. The model proposed here lacks language-specific assumptions (similar to the exemplar-based models). Comparing the performance of substance-free models with competing proposals and human behavior should result in a better understanding of what aspects of phonological grammar and acquisition are domain-specific (Section \ref{discussion}).

In the following, we first survey existing theories of phonological grammar and literature on computational approaches to  phonology (Section \ref{previous}).  In Section \ref{materials}, we present the model in \citep{donahue19} based on \cite{radford15} and provide acoustic and statistical analysis of  the training data. The network's outputs are first acoustically analyzed and described in Sections \ref{trag} and \ref{m12255}. In Section \ref{internalrep}, we present a technique for establishing the network's internal representations and test it with two generative tests. In Sections \ref{pvlv} and \ref{pvlv}, we analyze phonetic properties of the network's internal representations. Section \ref{discussion} compares the outputs of the model with L1 acquisition, speech impairments, and speech errors.
   
    \section{Previous work}
    
    \label{previous}

In the generative tradition, \emph{phonological grammar} derives surface phonetic outputs from phonological inputs \citep{ch68}.  For example, /p/ is an abstract unit that can surface (be realized) with variations at the phonetic level. English /p/ is realized as aspirated [\textipa{p\super h}] (produced with a puff of air) word-initially before stressed vowels, but as unaspirated plain [p] (without the puff of air) if [s] immediately precedes it. This distribution is completely predictable and derivable with a simple rule \citep{iverson95}, which is why the  English phoneme /p/ as an abstract mental unit is unspecified for aspiration (or absence thereof) in the underlying representation (/\textipa{"pIt}/ `pit'  and  /\textipa{"spIt}/ `spit'). The surface phonetic outputs after the phonological derivation had taken place are [\textipa{"p\textsuperscript{\textbf{h}}It}] with the aspiration and [\textipa{"spIt}] without the aspiration.

One of the main objectives of phonological theory is to explain how the grammar derives surface \emph{outputs}, i.e.~phonetic signals, from  \emph{inputs}, i.e.~phonemic representations. Two influential proposals have been in the center of this discussion, the rule-based approach and Optimality Theory. The first approach uses rewrite rules \citep{ch68} or finite state automata  \citep{heinz10,chandlee14} to derive outputs from inputs through derivation.  A connectionist approach called Optimality Theory \citep{prince9304} and related proposals such as Harmonic Grammar and Maximum Entropy  (MaxEnt) grammar \citep{legendre90,goldwater03,legendre06,w06,hayes08,pater09,wh13,white14,w17}, on the other hand, model phonological grammar as input-output pairing: the grammar chooses the most optimal output given an input. These models were heavily influenced by the early advances in neural network research \citep{alderete18,pater19}. Modeling linguistic data with neural networks has seen a rapid increase in the past few years (\citealt{alderete13,avcu17,kirov17,alderete18,mahalunkar18,weber18a,dupoux18,prickett19}, for cautionary notes, see \citealt{rawski19}). One of the promising implications of neural network modeling is the ability to test generalizations that the models produce without language-specific assumptions \citep{pater19}.

In opposition to the generative approaches, there exists a long tradition of usage-based models in phonology \citep{bybee99,silverman17} which diverges from the generative approaches in some crucial aspects. Exemplar models \citep{johnson97,p01,gahl06,johnson07,kaplan17}, for example, assume that phonetic representations are stored as experiences or exemplars. Grammatical behavior emerges as a consequence of generalization (or computation) over a cloud of exemplars \citep{johnson07,kaplan17}. In this framework, there is no direct need for a separate underlying representation from which the surface outputs are derived (or optimized). Several phenomena in phonetics and phonology have been successfully derived within this approach (for an overview, see \citealt{kaplan17}), and the framework allows phonology to be modeled computationally. The computational models often involve interacting agents learning some simplified phonetic properties (e.g.~\citealt{boer00,wedel06,kirby15}).

The majority of existing computational models in phonology (including finite state automata, the MaxEnt model and the existing neural network methods) model learning as symbol manipulation and operate with discrete units---either with completely abstract made-up units or with discrete units that feature some phonetic properties that can be approximated as phonemes. This means that either phonetic and phonological learning are modeled separately or one is assumed to have already been completed \citep{martin12,dupoux18}. This is true for both proposals that model phonological distributions or derivations \citep{alderete13,kirov17,futrell17,prickett19} and featural organizations \citep{faruqui16,silfverberg18}. The existing models also require strong assumptions about learning: underlying representations, for example, are pre-assumed and not inferred from data \citep{kirov17,prickett19}.

Most models in the subset of the proposals that operate with continuous phonetic data assume at least some level of abstraction and operate with already extracted features (e.g.~formant values) on limited ``toy'' data (e.g.~\citealt{p01,kirby15} for a discussion, see \citealt{dupoux18}).  \cite{guenther12}, \cite{guenther16} and \cite{oudeyer01,oudeyer02,oudeyer05,oudeyer06} propose models that use simple neural maps that are based on actual correlates of neurons involved in speech production in the human brain (based on various brain imaging techniques). Their models, however, do not operate with raw acoustic data (or require extraction of features in a highly abstract model of articulators; \citealt{oudeyer05,oudeyer06}), require a level of abstraction in the input to the model, and do not model phonological processes --- i.e.~allophonic distributions. Phonological learning in most of these proposals is thus modeled as if phonetic learning (or at least a subset of phonetic learning) has already taken place: the initial state already includes phonemic inventories, phonemes as discrete units, feature matrices that have already been learned, or extracted phonetic values. 

Prominent among the few models that operate with raw phonetic data are Gaussian mixture models for category-learning or phoneme extraction \citep{feldman19,lee12}.  \cite{feldman19} propose  a Dirichlet process Gaussian mixture model that learns categories from raw acoustic input in an unsupervised learning task. The model is trained on English and Japanese data and the authors show that the asymmetry in perceptual [l]$\sim$[r] distinction between English and Japanese falls out automatically from their model. The primary purpose of the proposal in \cite{feldman19} is modeling perception and categorization: they model  how a learner is able to categorize raw acoustic data into sets of discrete categorical units that have phonetic values (i.e.~phonemes). No phonological processes are modeled in the proposal. 

A number of earlier works in the connectionist approach that included basic neural network architectures to model  mapping from some simplified phonetic space to the discrete phonological space \citep{mccleland86,gaskell95,plaut99,kello03}. Input to most of these models is not raw acoustic data (except in \citealt{kello03}), but already extracted features. Learning in these models is also not unsupervised: the models come pre-specified with discretized phonetic or phonological units. None of the models are generative and do not model learning of phonological processes, but rather of classifying a simplified phonetic space with already available phonological elements.

 Recently, neural network models for unsupervised feature extraction have seen success in modeling acquisition of phonetic features from raw acoustic data  \citep{rasanen16,eloff19,shain19}. The model in \cite{shain19}, for example, is an autoencoder neural network that is trained on pre-segmented acoustic data. The model takes as input segmented acoustic data and outputs  values that can be correlated to phonological features. Learning is, however, not completely unsupervised as the network is trained on pre-segmented phones. \cite{thiollire15} similarly propose an architecture that extracts units from unsupervised speech data.  Other proposals for unsupervised acoustic analysis with neural network architecture are similarly primarily concerned with unsupervised feature extraction \citep{kamper15}.  These  proposals, however, do not model learning of phonological distributions, but only of feature representations, do not show a direct relationship between individual variables in the latent space and acoustic outputs (as in Section \ref{inter} and Figure \ref{interpolation2}), and crucially are not generative, meaning that the models do not output innovative data, but try to replicate the input as closely as possible (e.g.~in the autoencoder architecture). 
 
 As argued below, the model based on a Generative Adversarial Network (GAN) learns not only to generate innovative data that closely resemble human speech, but also learns internal representations that resemble phonological learning with unsupervised phonetic learning from raw acoustic data. Additionally, the model  is generative and outputs both the conditional allophonic distributions and  innovative data that can be compared to productive outputs in human speech acquisition.

\section{Materials}
\label{materials}

\subsection{The model: \cite{donahue19} based on \cite{radford15}}

\begin{figure}
\centering
\includegraphics[width=0.85\textwidth]{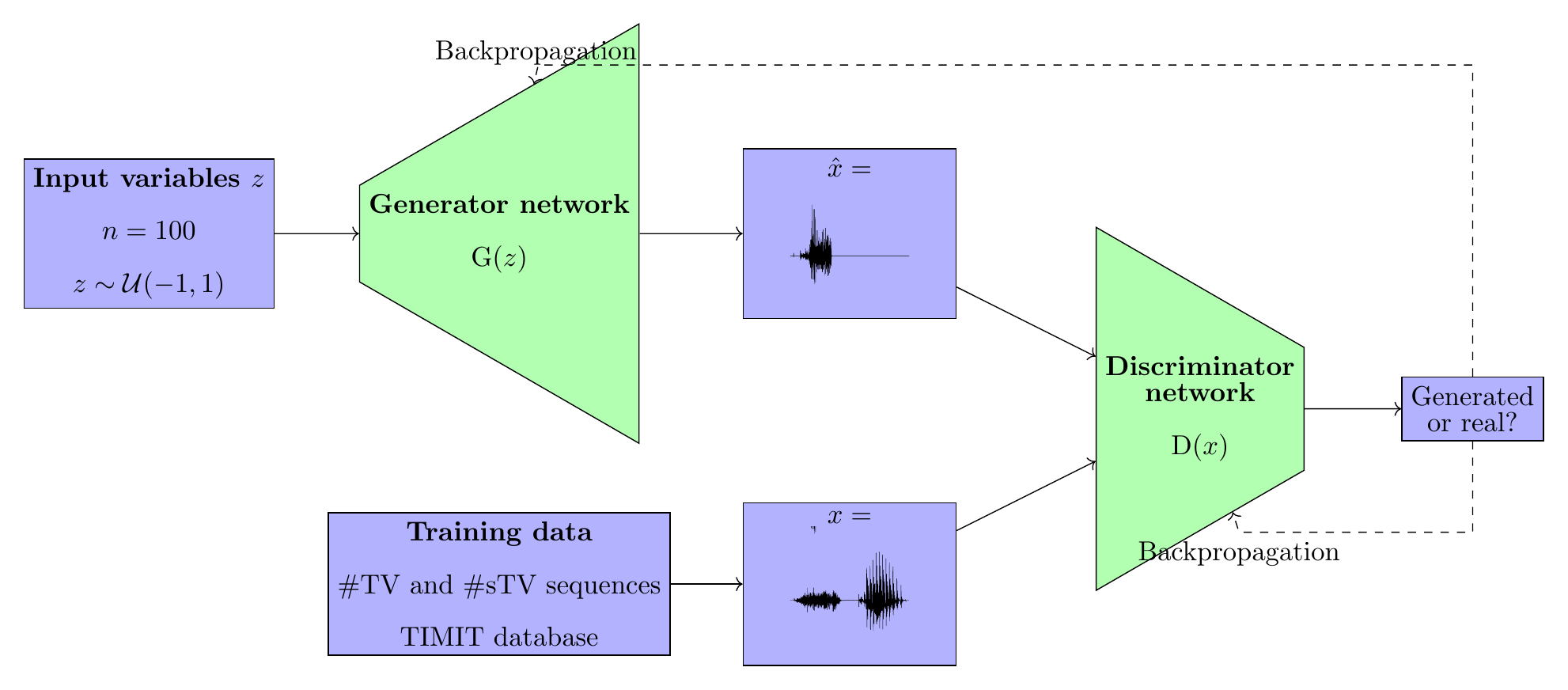}
\caption{\label{tikzflow} A diagram showing the Generative Adversarial architecture as proposed in \cite{goodfellow14,donahue19}, trained on data from the TIMIT database in this paper. }
\end{figure}

Generative Adversarial Networks, proposed by \cite{goodfellow14}, have seen a rapid expansion in a variety of tasks, including but not limited to computer vision and image generation \citep{radford15}. The main characteristic of GANs is the architecture that involves two networks: the Generator network and the Discriminator network  \citep{goodfellow14}. The Generator network is trained to generate data  from random noise, while the Discriminator is trained to distinguish real data from the outputs of the Generator network (Figure \ref{tikzflow}). The Generator is trained to generate data that maximizes the error rate of the Discriminator network.  The training results in a Generator (G) network that takes random noise as its input (e.g.~multiple variables with uniform distributions) and outputs data such that the Discriminator is inaccurate in distinguishing the generated from the real data (Figure \ref{tikzflow}). 

Applying the GAN architecture to time-series data such as a continuous speech stream poses several challenges. Recently, \citet{donahue19} proposed an implementation of a Deep Convolutional Generative Adversarial Network proposed by \citet{radford15} for audio data (WaveGAN); the model  along with the code in \cite{donahue19} were used for training in this paper. The model takes one-second long raw audio files as inputs, sampled at 16 kHz with 16-bit quantization. The audio files are converted into a vector and fed to the Discriminator network as real data. Instead of the two-dimensional $5\times5$ filters, the WaveGAN model uses one-dimensional $1\times25$ filters and larger upsampling. The main architecture is preserved as in DCGAN, except that an additional layer is introduced in order to generate longer samples \citep{donahue19}.  The Generator network takes as input $z$, a vector of one hundred uniformly distributed variables ($z\sim\mathcal{U}(-1,1)$) and outputs 16,384 data points, which constitutes the output audio signal. The network has five 1D convolutional layers \citep{donahue19}. The Discriminator network takes 16,384 data points (raw audio files) as its input and outputs a single value. The Discriminator's weights are updated five times per each update of the Generator. The initial GAN design as proposed by \cite{goodfellow14} trained the Discriminator network to distinguish real from generated data. Training such models, however, posed substantial challenges \citep{donahue19}. \cite{donahue19} implement the WGAN-GP strategy \citep{arjovsky17,gulrajani17}, which means that the Discriminator is trained ``as a function that assists in computing the Wasserstein distance'' \citep{donahue19}. The WaveGAN model \citep{donahue19} uses ReLU activation in all but the last layer for the Generator network, and Leaky ReLU in all layers in the Discriminator network (as recommended for DCGAN in \citealt{radford15}). For exact dimensions of each layer and other details of the model, see \cite{donahue19}.

\subsection{Training data}

The model was trained on  the allophonic distribution of voiceless stops in English. As already mentioned in Section \ref{intro}, voiceless stops /p, t, k/ surface as aspirated (produced with a puff of air) [\textipa{p\super h, t\super h, k\super h}] in English in word-initial position when immediately followed by a stressed vowel \citep{lisker84,iverson95,vaux02,vaux05,davis06}. If an alveolar sibilant [s] precedes the stop, however, the aspiration is blocked and the stop surfaces as unaspirated [p, t, k] \citep{vaux05}. A minimal pair illustrating this allophonic distribution is [\textipa{"p\super hIt}] `pit' vs.~[\textipa{"spIt}] `spit'. The most prominent phonetic correlate of this allophonic distribution is the difference in Voice Onset Time (VOT) duration \citep{abramson64,abramson17} between the  aspirated and unaspirated voiceless stops. VOT is the duration between the release of the stop ([p, t, k]) and the onset of periodic vibration in the following vowel. 

\begin{figure}
\centering
\includegraphics[width=0.49\textwidth]{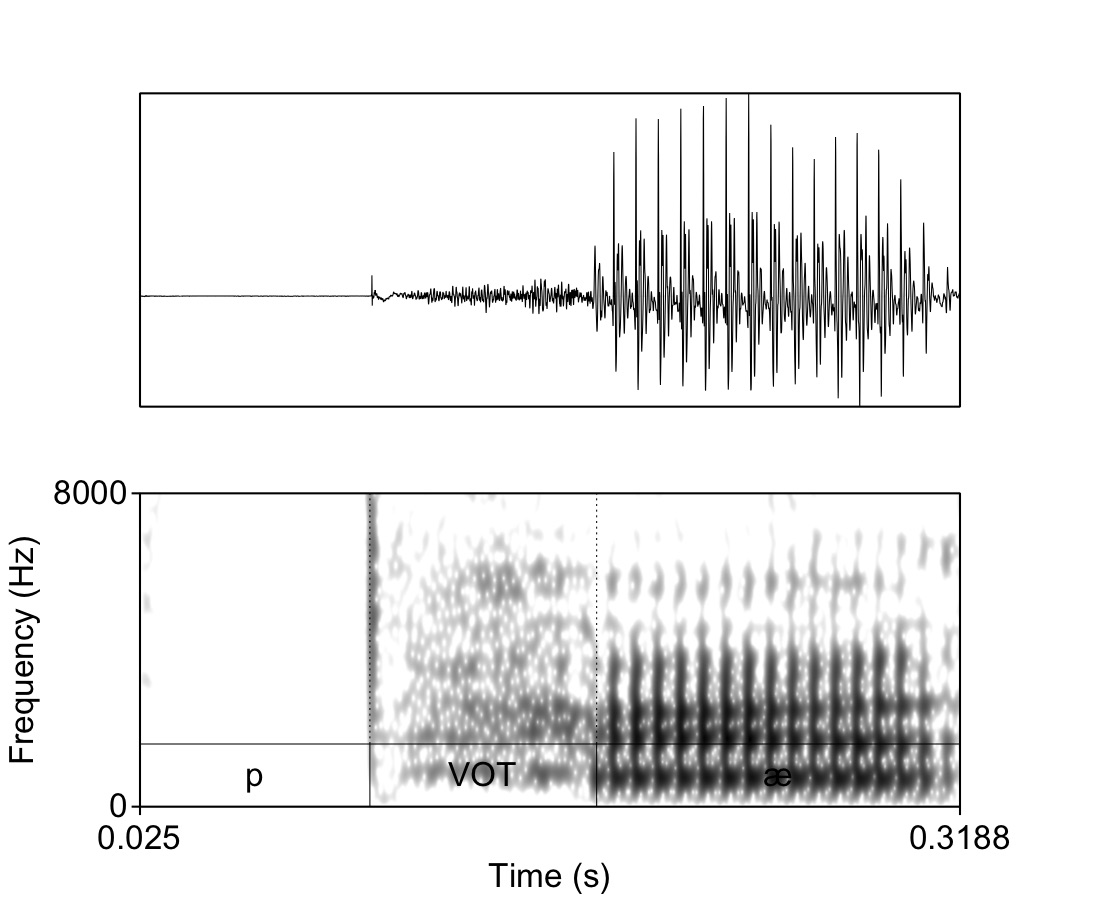}\includegraphics[width=0.49\textwidth]{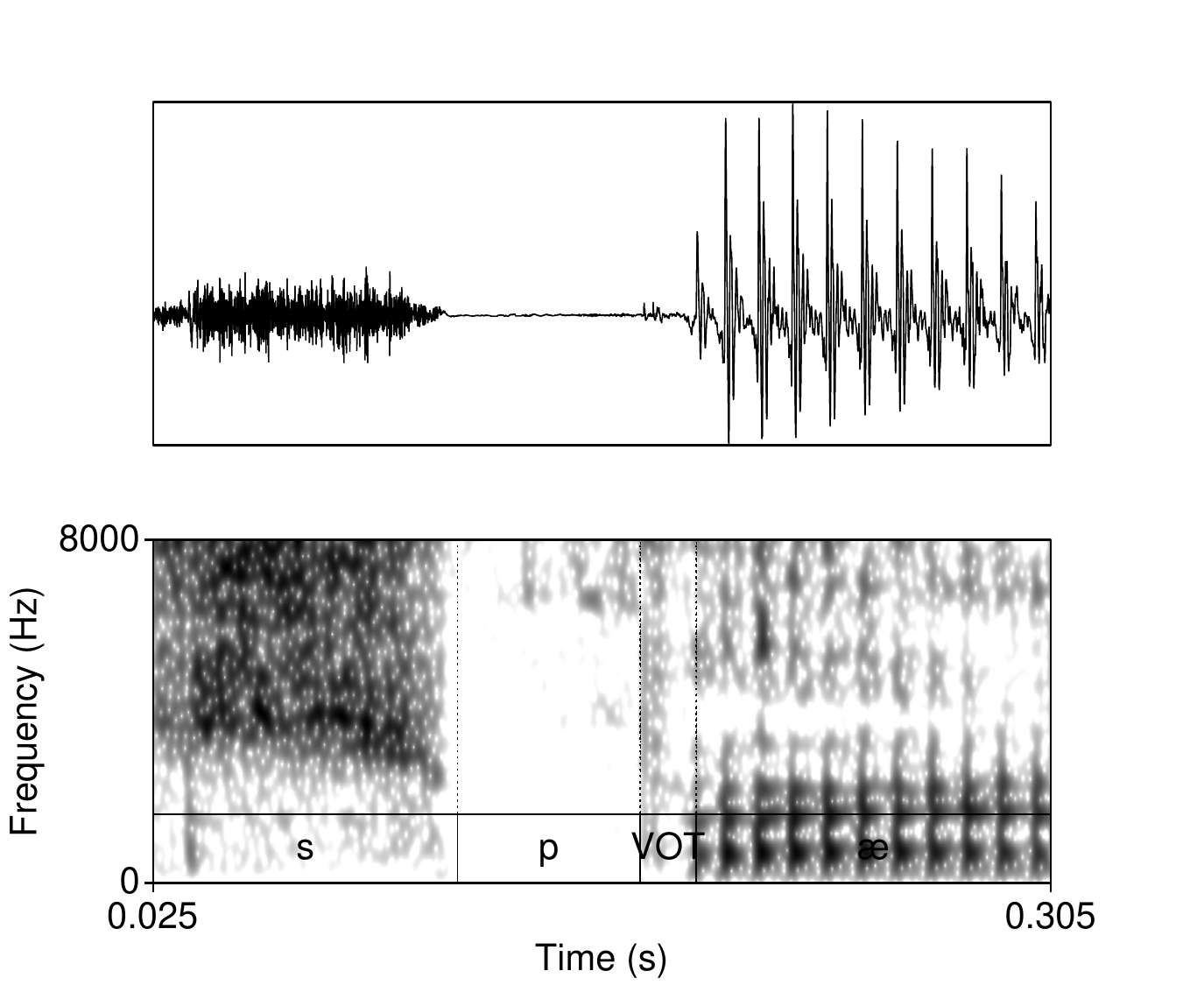}
\caption{\label{trainExam}Waveforms and spectrograms ($0-8000$ Hz) of [\textipa{p\super h\ae}] (left) and  [\textipa{sp\ae}] (right) illustrating  typical training data with annotations from TIMIT.  Only the raw audio data (in .wav format) were used in training. The annotation illustrates a substantially longer duration of VOT in word-initial stops when no [s] precedes. }
\end{figure}

The model was trained on data from the TIMIT database \citep{timit}.\footnote{\cite{donahue19} trained the model on the SC09 and TIMIT databases, but the results are not useful for modeling phonological learning, because the model is trained on a continuous speech stream and the generated sample fails to produce analyzable results for phonological purposes.} The corpus was chosen because it is one of the largest currently available hand-annotated speech corpora, the recording quality is relatively high, and the corpus features a relative high degree of variability. The database includes 6300 sentences, 10 sentences per 630 speakers from 8 major dialectal areas in the US \citep{timit}. The training data consist of 16-bit .wav files with 16 kHz sampling rate of word initial sequences of voiceless stops /p, t, k/ (= T) that were followed by a vowel (\#TV) and word initial sequences of /s/ + /p, t, k/, followed by a vowel (\#sTV). The training data includes 4,930  sequences with the structure \#TV (90.2\%) and 533 (9.8\%) sequences with the structure \#sTV (5,463 total). Figure \ref{trainExam} illustrates typical training data: raw audio files with speech data, but limited to two types of sequences, \#TV and \#sTV. Figure \ref{trainExam} also illustrates that the duration of VOT depends on a condition that is not immediately adjacent in phonetic terms: the absence/presence of [s] is interrupted from the VOT duration by a period of closure in the training data. That VOT is significantly shorter if T is preceded by [s]  in the training data is confirmed by a Gamma regression model: $\beta=-0.84,  t =-49.69, p<0.0001$ (for details, see Supplementary materials Section \ref{tdgr}).

\begin{figure}
\centering
\includegraphics[width=0.95\textwidth]{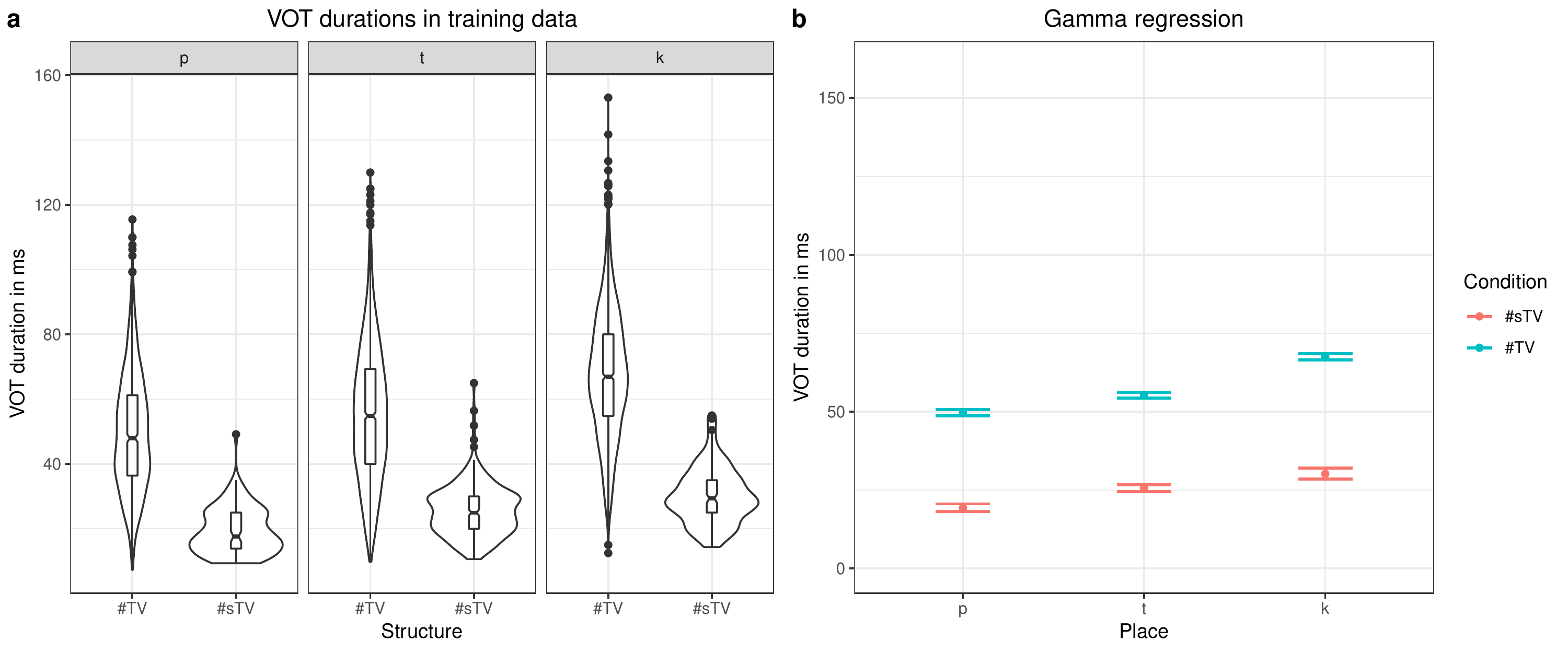} 
\caption{\label{violin1} \textbf{(a)} Violin plots with box-plots of durations in ms of VOT in the training data based on two conditions: when word-initial \#TV sequence is not preceded by [s] (\#sTV) and when it is preceded by [s] (\#sTV) accross the three places of articulation: [p], [t], [k]. \textbf{(b)} Fitted values of VOT durations with 95\% confidence intervals from the Gamma (with log-link) regression model in Table \ref{timitlm}.}
\end{figure}

Both stressed and unstressed vowels are included in the training data. Including both stressed and unstressed vowels is desirable, as this condition crucially complicates learning and makes the task for the model more challenging as well as more realistic. Aspiration is less prominent in word-initial stops not followed by a stressed vowel. This means that in the condition \#TV, the stop will be either fully aspirated (if followed by a stressed vowel) or unaspirated (if followed by an unstressed vowel). Violin plots in Figure  \ref{violin1} illustrate that aspiration of stops before an unstressed vowel can be as short as in the \#sTV condition.  In the \#sTV condition, the stop is never aspirated. Learning of two conditions is more complex if the dependent variable in one condition can range across the variable in the other condition.

\begin{table}
\centering
\begin{tabular}{ccrrrrr}
  \hline
 \textbf{Structure} & \textbf{Place} & \textbf{VOT} & \textbf{SD} & \textbf{Lowest} & \textbf{Highest}& \textbf{Count}  \\ 
  \hline
\multirow{3}{*}{\#TV} & p & 49.6 & 18.0 & 7.3 & 115.5&1018 \\ 
  & t & 55.2 & 20.7 & 9.8 & 130.0&1799 \\ 
  & k & 67.5 & 19.5 & 12.5 & 153.1&2112 \\ \hline
\multirow{3}{*}{\#sTV}  & p & 19.4 & 7.1 & 9.4 & 49.2&115 \\ 
& t & 25.6 & 7.9 & 10.6 & 65.0&288 \\ 
 & k & 30.1 & 8.6 & 14.4 & 55.0 &130\\ 
   \hline
\end{tabular}
\caption{\label{durations}Raw VOT durations in ms for the training data with SD and Range.}
\end{table}

\begin{figure}
\centering
\includegraphics[width=0.8\textwidth]{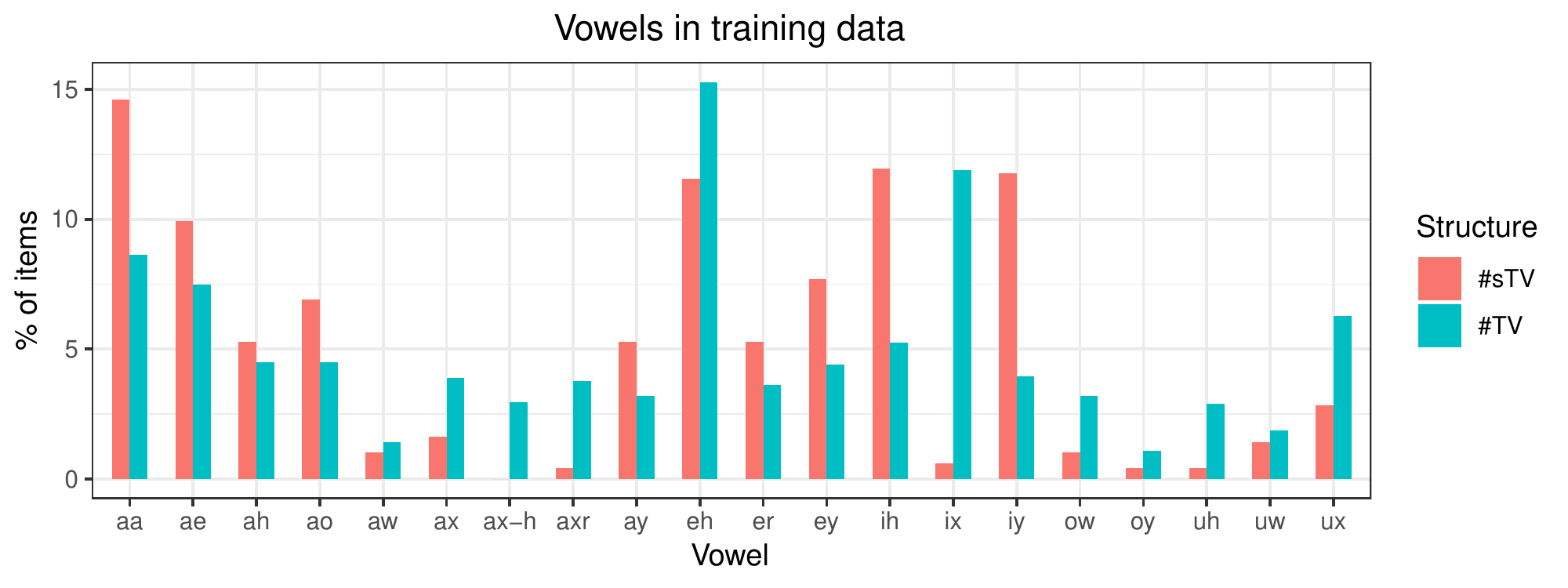}
\caption{\label{timitProc} Distribution of training items according to vowel identity as described in TIMIT in ARPABET, where aa = \textipa{A}, ae = \textipa{\ae}, ah = \textipa{\textturnv}, ao = \textipa{O}, aw = \textipa{aU}, ax = \textipa{@}, ax-h = \textipa{\r*{@}}, axr = \textipa{\textrhookschwa}, ay = \textipa{aI}, eh = \textipa{E}, er = \textipa{\textrhookrevepsilon}, ey = \textipa{eI}, ih = \textipa{I}, ix = \textipa{1}, iy = \textipa{i}, ow = \textipa{oU}, oy = \textipa{oI}, uh = \textipa{U}, uw = \textipa{u}, ux = \textipa{0} in the International Phonetic Alphabet. }
\end{figure}

The training data is not completely naturalistic: \#TV and \#sTV sequences are sliced from continuous speech data. This, however, has a desirable effect. The primary purpose of this paper is to test whether a GAN model can learn an allophonic distribution from data that consists of raw acoustic inputs. If the entire lexicon was included in the training data, the distribution of VOT duration could be conditioned on some other distribution, not the one this paper is predominately interested in: the presence or absence of [s]. It is thus less likely that the distribution of VOT duration across the main condition of interest, the presence of [s], is conditioned on some other unwanted factor in the model precisely because of the balanced design of the training data. The only condition that can potentially influence learning is the distribution of vowels across the two conditions. Figure \ref{timitProc}, however, shows that vowels are relatively equally distributed across the two conditions, which means that vowel identity likely does not influence the outcomes substantially. Finally, vowel duration (or the equivalent of speech rate in real data) and identity are not controlled for in the present experiment. To control for vowel duration, VOT duration would have to be modeled as a proportion of the following vowel duration. Several confounds that are not easy to address would be introduced, the main of which is that vowel identification is problematic for generated inputs with fewer training steps. Because the primary interest of the experiment is the difference in VOT durations between two groups (the presence and absence of [s]) and substantial differences in vowel durations (or speech rate) between the two groups are not expected, we do not anticipate the results to be substantially influenced by speech rate.

\section{Experiment}
\label{experiment}

\subsection{Training and generation}
\label{trag}

 The purpose of this paper is to model phonetic and phonological learning. For this reason, the data was generated and examined at different points as the Generator network was in the process of  being trained. For the purpose of modeling learning, it is more informative to probe the networks with fewer training steps, which allows a comparison between the model's outputs and L1 acquisition (Section \ref{parallelshuman}). Outputs of the network are analyzed  after 12,255 steps (Section \ref{m12255}). The number of steps was chosen  as a compromise between quality of output data and the number of epochs in the training. Establishing the number of training steps at which an effective acoustic analysis can be performed is at this point somewhat arbitrary. We generated outputs of the Generator model trained after 1474, 4075, 6759, 9367, and 12,255 steps and manually inspected them. The model trained after 12,255 steps was considered the first that allowed a reliable acoustic analysis based on quality of the generated outputs. It would be informative to test how accuracy of labeled data improves with training steps, but this is left for future work.  The model was trained on a single NVIDIA K80 GPU. The network was trained at an approximate pace of 40 steps per 300 s. In Section \ref{m12255}, we present measurements of VOT durations in the \#sTV and \#TV conditions in the Generated outputs and discuss linguistically interpretable innovative outputs that violate the training data. In Section \ref{regression}, we propose a technique for recovering the Generator network's internal representations; in Section \ref{inter} we illustrate that manipulating these variables have a phonetically meaningful effect in the output data.

\subsection{\label{m12255}VOT duration}

\begin{figure}
\centering

\includegraphics[width=0.4\textwidth]{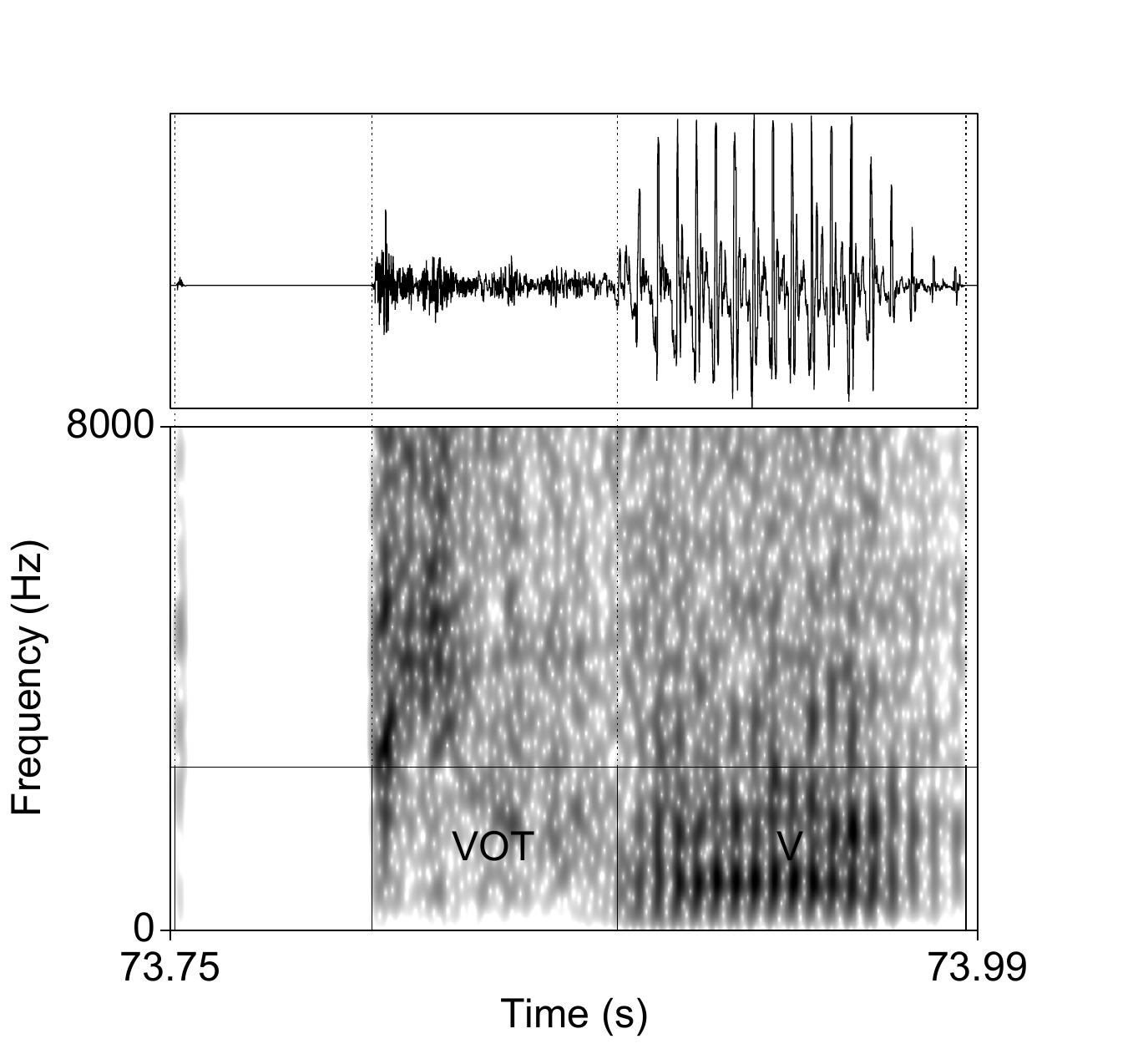}
\includegraphics[width=0.4\textwidth]{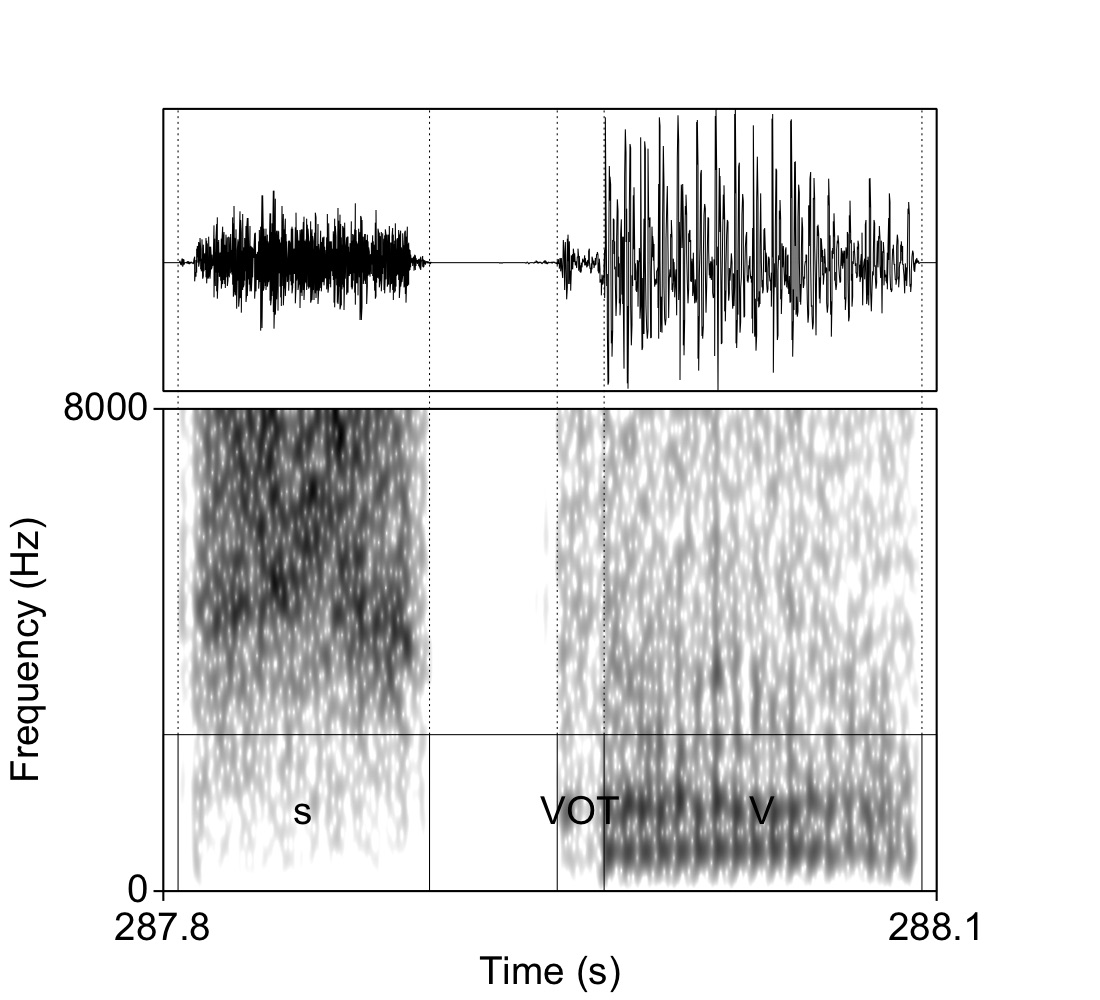}

\caption{\label{12275a}Waveforms and spectrograms (0--8,000 Hz) of typical generated samples of \#TV (left) and \#sTV (right) sequences from a Generator trained after 12,255 steps. }
\end{figure}

The Generator network after 12,255 steps ($\sim$ 716 epochs) generates acoustic data that appear close to actual speech data. Figure \ref{12275a} illustrates a typical generated sample of \#TV (left) and \#sTV (right) structures with a substantial difference in VOT durations.

\begin{figure}
\centering
\includegraphics[width=1\textwidth]{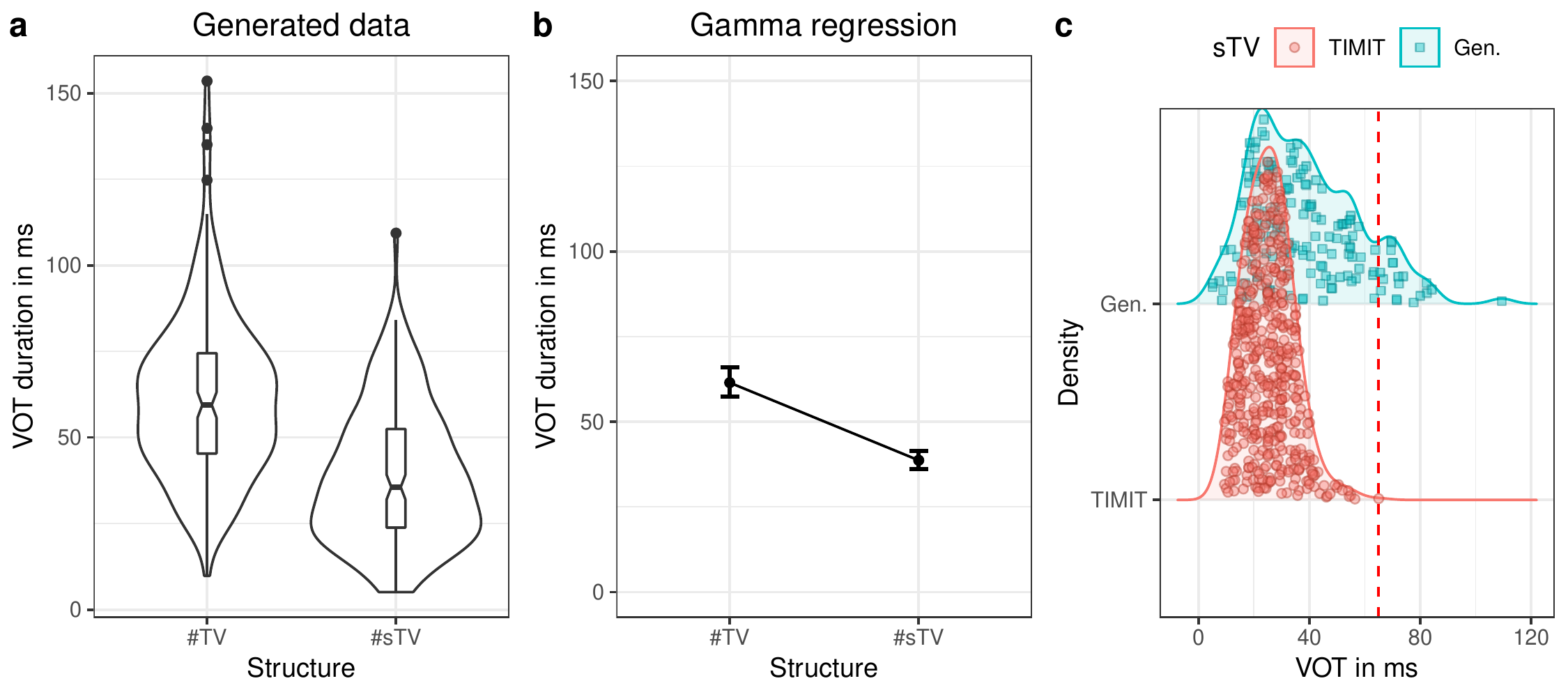} 
\caption{\label{violin2} \textbf{(a)} Violin plots with box-plots of durations in ms of VOT in the generated data based on two conditions: when word-initial \#TV sequence is not preceded by [s] (\#sTV) and when it is preceded by [s] (\#sTV). \textbf{(b)} Estimates of VOT duration based on fitted values with 95\% confidence intervals across two conditions, \#TV and \#sTV in the generated data. \textbf{(c)} Duration (in ms) of VOT in the \#sTV condition that compares the TIMIT training data (red circles) and generated outputs (green squares). The dashed line represents the longest VOT duration in the training data. The figure illustrates the proportion of outputs that violate the training data. }
\end{figure}

 To test whether the Generator learns the conditional distribution of VOT duration, 2000 samples were generated and manually inspected. First, VOT duration was manually annotated in all \#sTV sequences. There were altogether 156 such sequences. To perform significance testing on a similar sample size, the first 158 sequences of the \#TV structure were also annotated for  VOT duration. VOT was measured from the release of closure to the onset of periodic vibration with clear formant structure. Altogether 314 generated samples were thus annotated. Only samples with structure that resembles real acoustic outputs and for which VOT could be determined were annotated. The proportion of inputs for which a clear \#sTV or \#TV sequence was not recognizable is relatively small: in only 8 of the first 175 annotated outputs (4.6\%) was it not possible to estimate the VOT duration or whether the sequence is of the \#TV or a \#sTV structure. 
 Figure \ref{violin2} shows the raw distribution of VOT durations in the generated samples that closely resembles the distribution in the training data (Figure \ref{violin1}).

The results suggest that the network does learn the allophonic distribution: VOT duration is significantly shorter in the \#sTV condition ($\beta=-2.79, t= -78.34, p<0.0001$; for details of the statistical model, see Supplementary materials Section \ref{gammaregapp}). Figure \ref{violin2} illustrates estimates of VOT duration across the two conditions with 95\% confidence intervals. The model, however, shows clear traces that the learning is imperfect and that the generator network fails to learn the distribution \emph{categorically}. This is strongly suggested by the fact that VOT durations are substantially longer in the generated data compared to the training data. The difference in means between the \#TV and \#sTV conditions in the training data is 32.35 ms, while in the generated data the difference is 22.52 ms. The ratio between the two conditions in the training data is 2.34, while the generated data's ratio is 1.59. 

Another aspect of generated data that also strongly suggests the learning is imperfect is the fact that the longest VOT durations in the \#sTV condition in the generated data are substantially longer than the longest VOT durations in the training data,  where the longest duration reaches 65 ms (see Table \ref{durations} and Figure \ref{violin1}). VOT in the generated data is in 19 out of 156 total  \#sTV sequences (or 12.2\%) longer than 65.5 ms, the longest VOT in the training data.  The longest three VOT durations in \#sTV sequences are, for example, 109.35 ms, 84.17 ms, 82.37 ms.

\begin{figure}
\centering

\includegraphics[width=0.4\textwidth]{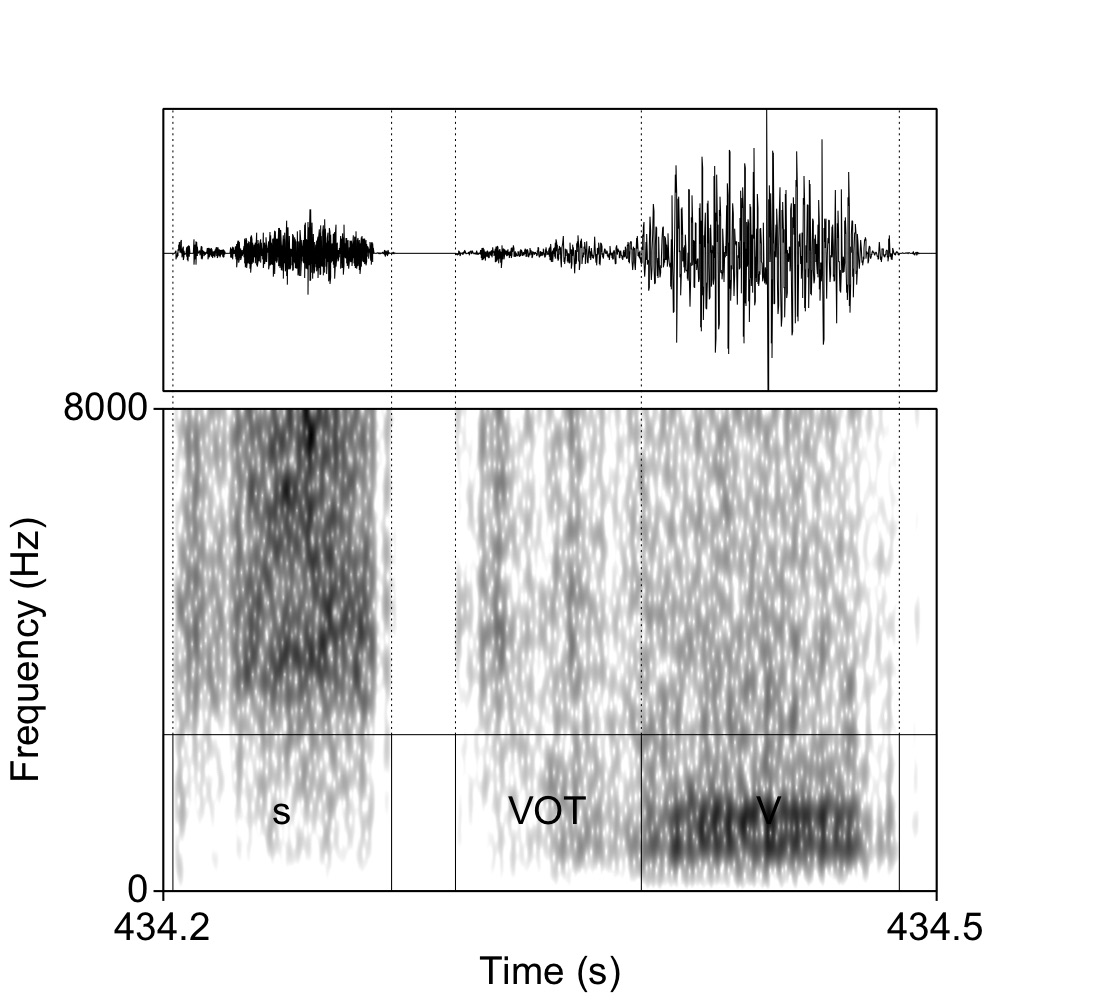}
\includegraphics[width=0.4\textwidth]{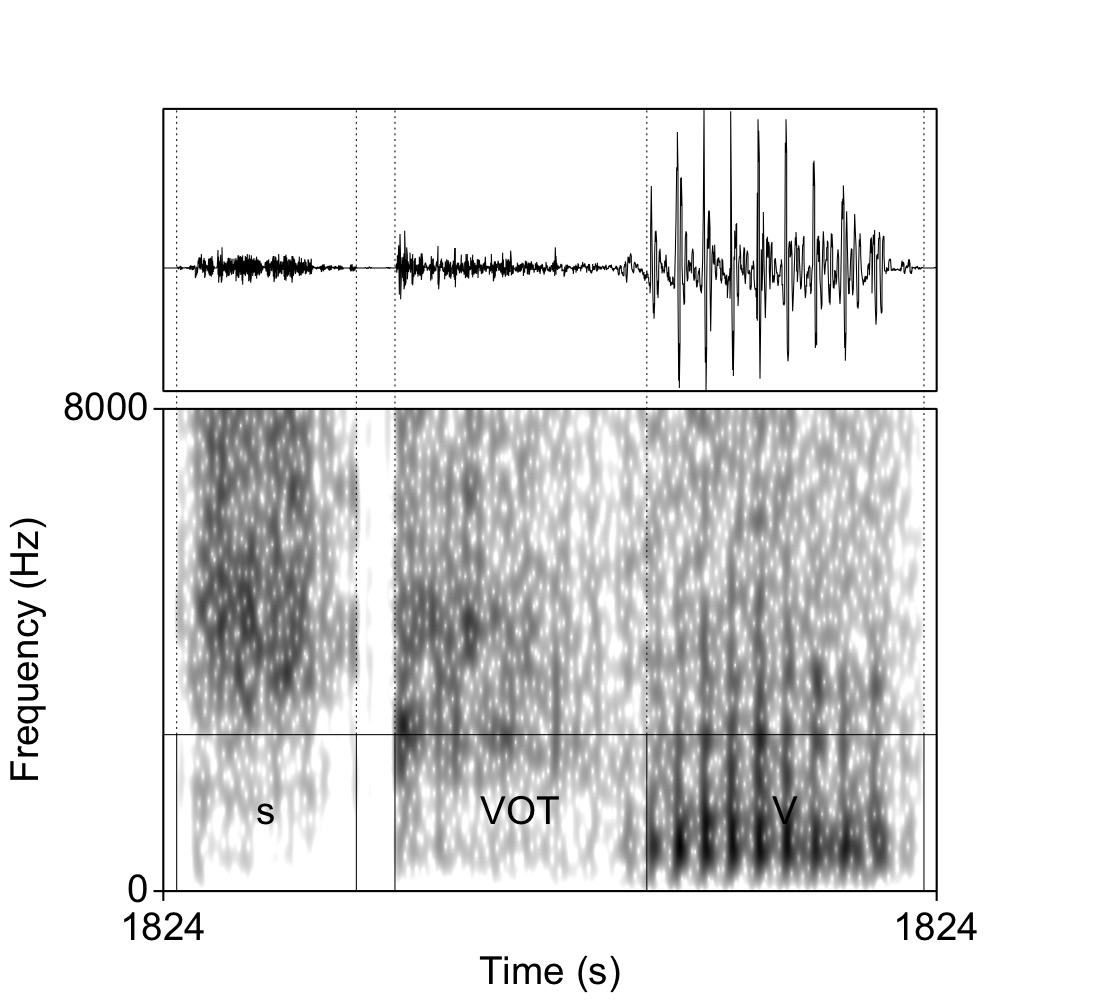}

\caption{\label{dolgVOT12255}Waveforms and spectrograms (0--8000 Hz) of two generated outputs of \#sTV sequences in which the stop has longer VOT than any VOT in \#sTV condition in the training data.}
\end{figure}

 This generalization holds also in proportional terms. To control for the overall duration of segments in the output, we measure ratio of VOT duration and duration of the preceding [s]  (i.e.~thus controlling for ``speech rate'').  The highest ratio between the VOT duration and the duration of preceding [s] ($\frac{\text{VOT}}{[\text{s}]}$) in the training data is 0.77\footnote{The TIMIT annotations would yield a ratio of 1.17, but the token was annotated by the author and the ratio appears much smaller. In any case, even with TIMIT's annotation, the ratio with value of 1.91 in the generated data is still substantially higher than the 1.17.}, which appears in an acoustically very different token compared to the generated outputs.  The ratio in all other tokens in the training data are even lower, below 0.69. 
Several values of the ratios between VOT and [s] duration in the training data are substantially higher compared to  the training data. In the three outputs with longest absolute duration of VOT, the ratios are 1.91, 1.40, and 0.89. Other high ratios measured include, for example,  1.79, 1.72, 1.60, 1.50, 1.46. Figure \ref{dolgVOT12255} shows two such cases. It is clear that the generator fails to reproduce the conditioned durational distribution from the training data in these particular cases. In other words, while the Generator learns to output significantly shorter VOT durations when [s] is present in the output, it occasionally (in approximately 12.2\% of cases) fails to observe this generalization and outputs a long VOT in the \#sTV condition which is longer than any VOT duration in the \#sTV condition in the training data.   As will be argued in Section \ref{parallelshuman}, the outcomes of this imperfect learning closely resemble L1 acquisition.

\begin{figure}
\centering
\includegraphics[width=0.8\textwidth]{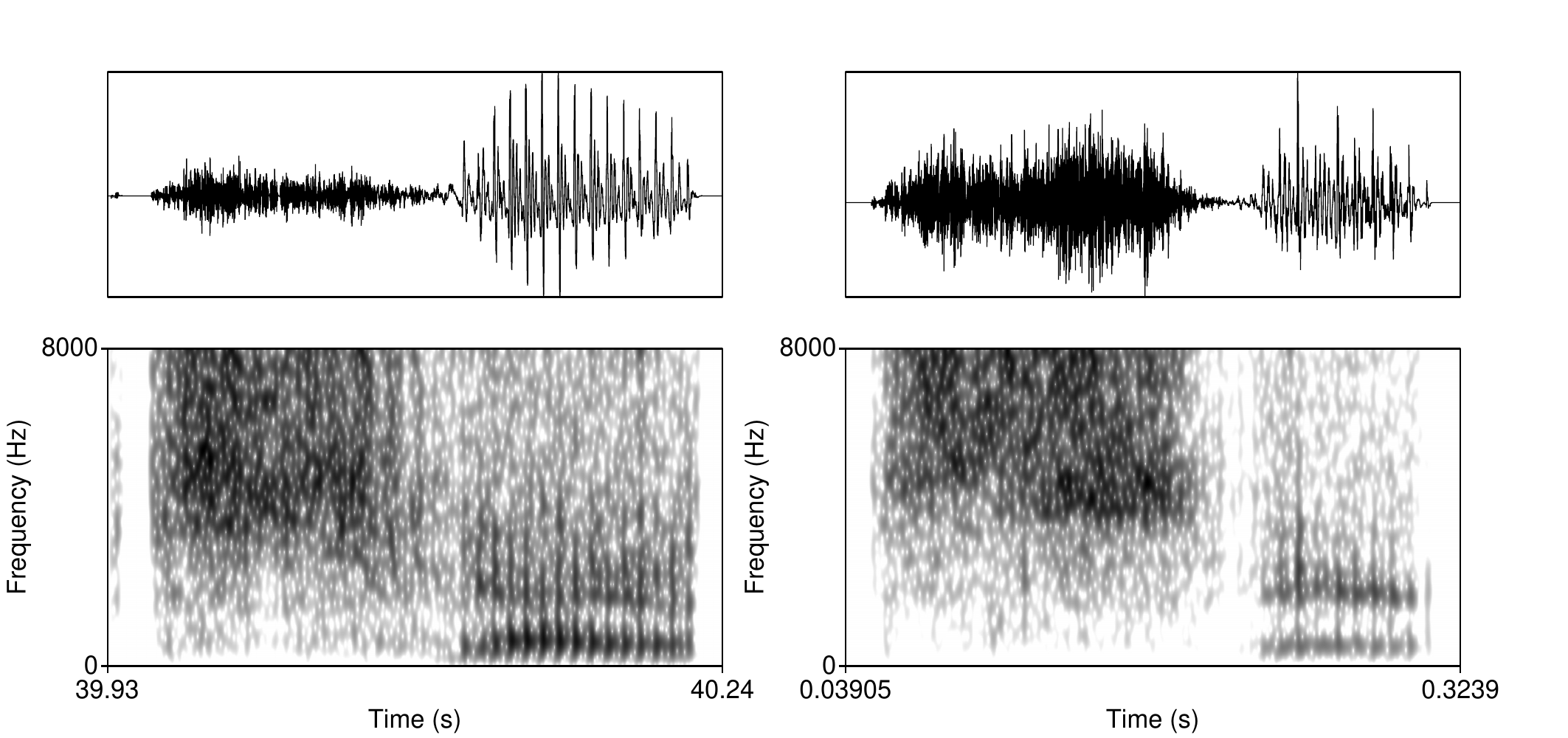}
\caption{\label{lepS}Waveforms and spectrograms (0--8000 Hz) of two generated outputs of the shape \#sV sequences. The  sample on the left was generated after 16,715 steps.}
\end{figure}

Longer VOT duration in the \#sTV condition in the generated data compared to training data is not the only violation of the training data that the Generator outputs and that resembles linguistic behavior in humans. Among approximately 3000 generated samples analyzed, we observe generated outputs that feature only frication noise of [s] and periodic vibration of the following vowel, but lack stop elements completely (e.g.~closure and release of the stop). In other words, the generator occasionally outputs a linguistically valid and innovative \#sV sequence for which no evidence was available in the training data. Such innovative sequences in which the segments are omitted or inserted are rare compared to innovative outputs with longer VOT -- approximately two per 3000 inspected cases (but the overall rate of outputs that are acoustically difficult to analyze is also small:  4.6\%). All sequences containing [s] from the training data were manually inspected by the author and none of them contain a \#sV sequence without a period of closure and VOT.  The minimal duration of closure in \#sTV sequences in the training data is 9.2 ms, and the minimal duration of VOT is 9.4 ms. Aspiration noise in stops that resembles frication of [s] and homorganic sequences of [s] followed by an alveolar stop [t] (\#stV) are occasionally acoustically similar to the sequence without the stop (\#sV) due to similar articulatory positions or because frication noise from [s] carries onto the homorganic alveolar closure which can be very short. Such data points in the training data can serve as the basis for the innovative output \#sV. However, there is a clear fall and a second rise of noise amplitude after the release of the stop in \#stV sequences.  Figure \ref{lepS} shows two cases of the Generator network outputting an innovative \#sV sequence without any stop-like fall of the amplitude, for which no evidence exists in the training data. 

\begin{figure}
\centering
\includegraphics[width=0.4\textwidth]{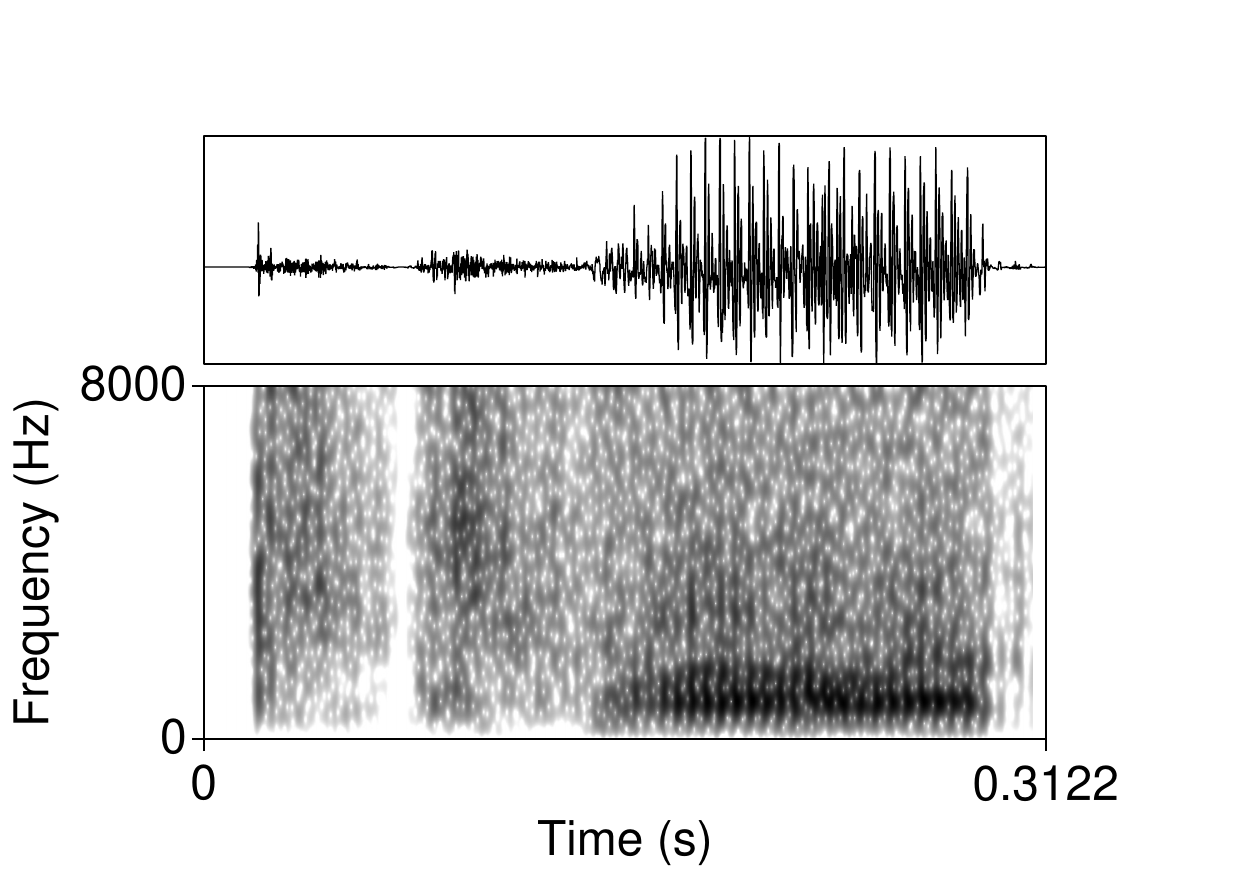}\includegraphics[width=0.4\textwidth]{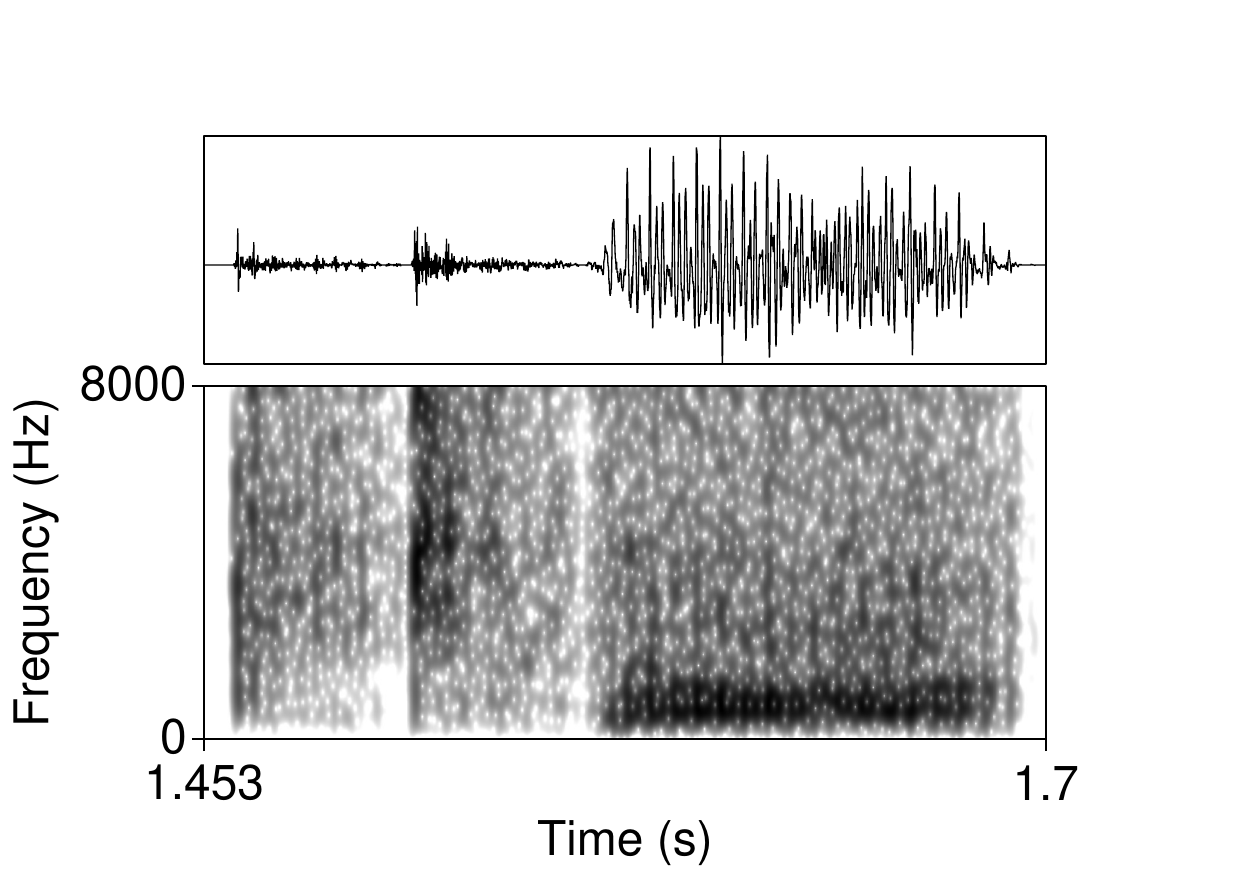}
\caption{\label{pta}Waveforms and spectrograms (0--8000 Hz) of two generated outputs of the shape \#TTV sequences.}
\end{figure}

Similarly, the Generator occasionally outputs a sequence with two stops and a vowel (\#TTV).  One potential source of such innovative sequences might be residual noise that is sometimes present during the period of closure in the TIMIT database. However, residual noise in the training data differs substantially from a clear aspiration noise in the generated \#TTV sequences.
 Figure \ref{pta} illustrates two generated examples in which the vocalic period is preceded by two bursts, two periods of aspiration and a short period of silence between the aspiration noise of the first consonant and the burst of the second consonant that corresponds to closure of the second stop.\footnote{For evidence that units smaller than segments are phonologically relevant, see \cite{inkelas17} and literature therein.} Spectrograms show the distribution of energy differs across the two bursts and aspiration noises, suggesting that the output represents a heterogranic cluster [pt] followed by a vowel.  

Measuring overfitting is a substantial problem for Generative Adversarial Networks with no consensus on the most appropriate quantitative approach to the problem \citep{goodfellow14,radford15}. The risk with overfitting in a GAN architecture is that the Generator network would learn to fully replicate the input.\footnote{In general, GANs do not overfit \citep{adlam19,donahue19}  as is suggested by our data. Even if overfitting did occur, it would result from training a Generator without a direct access to the training data (unlike in the autoencoder models, where the input training data and outputs are directly connected).} The best evidence against overfitting is precisely the fact that the Generator network outputs samples that substantially violate output distributions (Figures \ref{dolgVOT12255}, \ref{lepS}, and \ref{pta}).\footnote{\cite{donahue19} test overfitting on models trained with a substantially higher number of steps (200,000) compared to our model (12,255) and presents evidence that GAN models trained on audio data do not overfit even with substantially higher number of training steps.}

\subsection{\label{internalrep}Establishing internal representations}

Establishing what and how neural networks learn is a challenging task \citep{lillicrap19}. Exploration of latent space in the GAN architecture has been performed before \citep{radford15,lipton17,donahue17}, but to the author's knowledge, most previous work did not focus on discovering meaningful values (phonetic correlates) of each variable and have not fully used the potential to extend those variables to values well outside the training range (15 or 25). Below, we propose a technique for uncovering dependencies between the network's latent space and generated data based on logistic regression. We first use regression estimates to identify variables with a desired effect on the output by correlating the outputs of the Generator with its corresponding input variables that are uniformly distributed with an interval $(-1, 1)$ during training. We then Generate outputs by setting the identified latent variables to values well beyond the training range (to 4.5, 15, or 25). This method has the potential to reveal the underlying values of latent variables and  shed light on the network's internal representations. Using the proposed technique, we can estimate how the network learns to map from latent space to phonetically and phonologically meaningful units in the generated data.

To identify dependencies between the latent space and generated data,  we correlate annotations of the output data with the variables in the latent space (in Section \ref{regression}). As a starting point, we choose to identify correlates of the most prominent feature in the training data: the presence or absence of [s]. Any number of other phonetic features can be correlated with this approach (for future directions, see Section \ref{conclusion}); applying this technique to other features and other alternations should yield a better understanding of the network's learning mechanisms. Focusing on more than the chosen feature, however,  is beyond the scope of this paper.

\subsubsection{Regression}
\label{regression}

First, 3,800 outputs from the Generator network were generated and manually annotated for the presence or absence of [s]. 271 outputs (7.13\%) were annotated as involving a segment [s] which is similar to the percentage of data points with [s] in the training data (9.8\%). Frication that resembled [s]-like aspiration noise after the alveolar stop and before high vowels was not annotated as including [s].\footnote{It is possible that some outputs were mislabeled, but the probability is low and the magnitude of mislabeled data would be minimal enough not to influence the results. The author manually inspected spectrograms of all generated data.} Innovative outputs such as an \#[s] without the following vowel or \#sV sequences were annotated as including an [s].

The annotated data together with values of latent variables for each generated sample ($z$) were fit to a logistic regression generalized additive model (using the \emph{mgcv} package; \citealt{mgcv} in \citealt{r}) with the presence or absence of [s] as the dependent variable (binomial distribution of successes and failures) and smooth terms of latent variables ($z$) as predictors of interest (estimated as penalized thin plate regression splines; \citealt{mgcv}). Generalized additive models were chosen in order to avoid assumptions of linearity: it is possible that latent variables are not linearly correlated with features of interest in the output of the Generator network. The initial full model (\textsc{Full}) includes smooths for all 100 variables in the latent space that are uniformly distributed  within the interval $(-1,1)$ as predictors. 

The models explored here do not serve for hypothesis testing, but for exploratory purposes: to  identify variables, the effects of which are tested with two independent generative tests (see Sections \ref{Generativetest1} and \ref{gen2}). For this reason, several techniques to reduce the number of variables in the model with different shrinkage techniques are explored and compared: the latent variables for further analysis are then chosen based on combined results of different exploratory models. 

First, we refit the model with modified smoothing penalty (\textsc{Modified}), which allows shrinkage of the whole term \citep{mgcv}. Second, we refit the model with original smoothing penalty (\textsc{Select}), but with an additional penalty for each term if all smoothing parameters tend to infinity \citep{mgcv}. Finally, we identify non-significant terms by Wald test for each term (using \textit{anova.gam()} with $\alpha=0.05$) and manually remove them from the model (\textsc{Excluded}). 38 predictors are thus removed.

\begin{figure}
\centering
\includegraphics[width=0.99\textwidth]{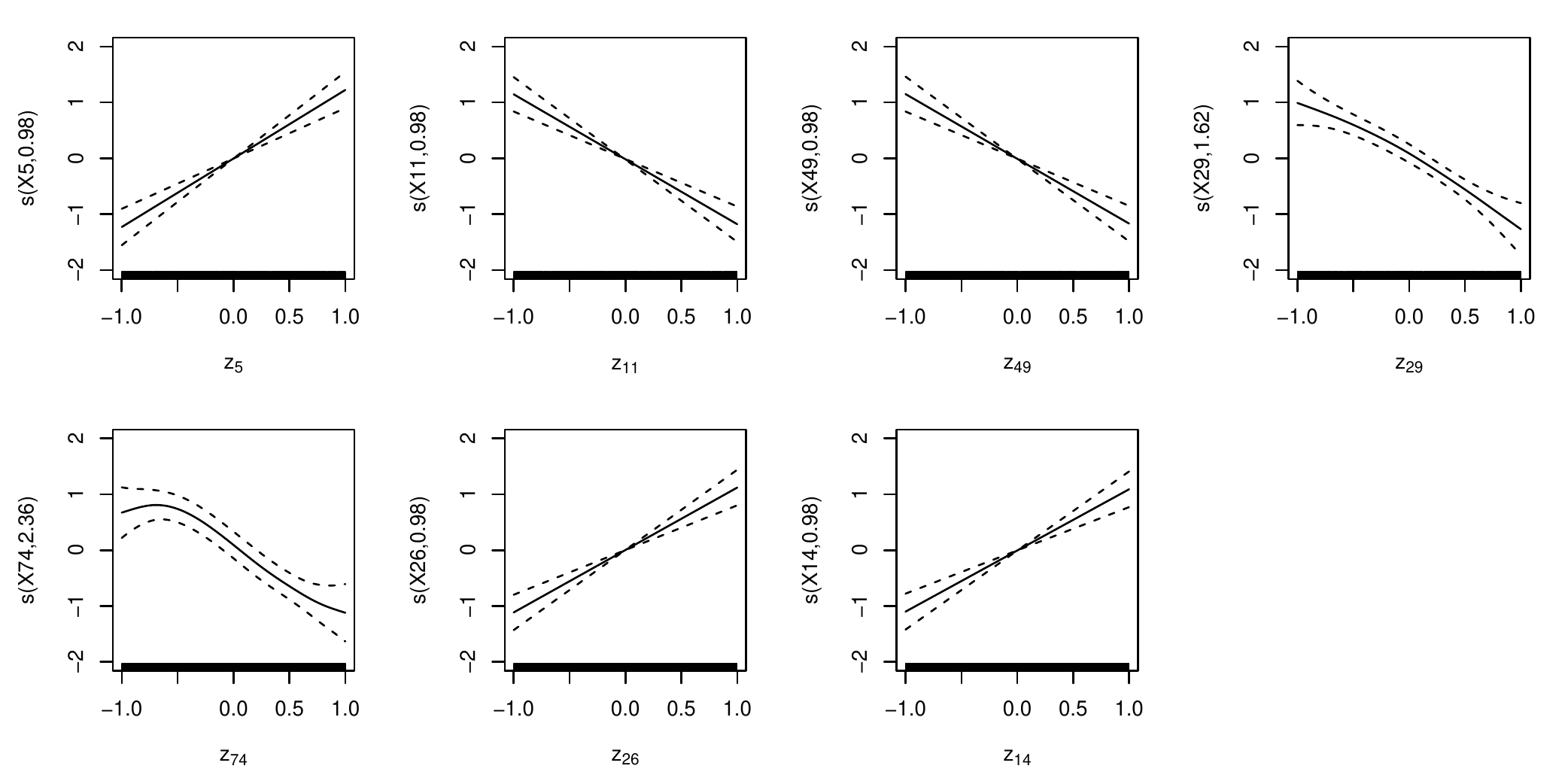}
\caption{\label{gam} Plots of seven smooth terms with highest $\chi^2$ values in a generalized additive logistic regression model with all 100 latent variables ($z$) as predictors, estimated with penalty for each term (\textsc{Select}). Many of the predictors show linear correlation, which is why a linear logistic regression outputs similar estimates.}
\end{figure}

The estimated smooths  appear mostly linear (Figure \ref{gam}). We also fit the data  to a linear logistic regression model (\textsc{Linear}) with all 100 predictors. To reduce the number of predictors, another model is fit (\textsc{Linear excluded}) with those predictors removed that do not improve fit (based on the AIC criterion when each predictor is removed from the full model). 23 predictors are thus removed. The advantage of the linear model is that predictors are parametrically estimated.\footnote{It would be possible to estimate smooth terms for only a subset of predictors, but such a model is unlikely to yield different results.} 

While the number of predictors in the models is high even after shrinkage or exclusion, there is little multicollinearity in the data as the 100 variables are randomly sampled for each generation. The highest Variance Inflation Factor in the linear logistic regression models (\textsc{Linear} and \textsc{Linear excluded}) estimated with the \emph{vif()} function (in the \emph{car} package; \cite{car}) is 1.287. All concurvity estimates in the non-linear models are below 0.3 (using \emph{concurvity()} in \citealt{mgcv}). While the number of successes per predictor is relatively low, it is unlikely that more data would yield substantially different results (as will be shown in Sections \ref{Generativetest1} and \ref{gen2}, the model successfully identifies those values that have direct phonetic correlates in the generated data).  

Six models are thus fit in an exploratory method to identify variables in the latent space that predict the presence of [s] in generated outputs. Table \ref{aics} lists AIC for each model. The \textsc{LinearExcluded} model has the lowest AIC score.  All six models, however, yield similar results. For further tests based on Lasso regression and Random Forest models that also yield similar results, see Supplementary material Section \ref{lrrf}.

\begin{table}
\centering
\begin{tabular}{rrr}
  \hline
 & df & AIC \\ 
  \hline
  
\textsc{Full} & 108.94 & 1018.38 \\ 
\textsc{Modified}& 88.06 & 1031.03 \\ 
\textsc{Excluded}& 71.51 & 1008.20 \\ 

\textsc{Linear}& 101.00 & 1036.04 \\ 
\textsc{Linear excluded} & 78.00 & 1007.06 \\ 
   \hline

\end{tabular}

   \caption{\label{aics}AIC values of five fitted models with corresponding degrees of freedom (df), fitted with Maximum Likelihood. AIC of \textsc{Select} is not listed because it was not fitted with ML; AIC of \textsc{Select} fitted with REML is, however, similar to \textsc{Excluded} (=1,008.46 vs.~1008.54).}
\end{table}

\begin{figure}
\centering
\includegraphics[width=0.8\textwidth]{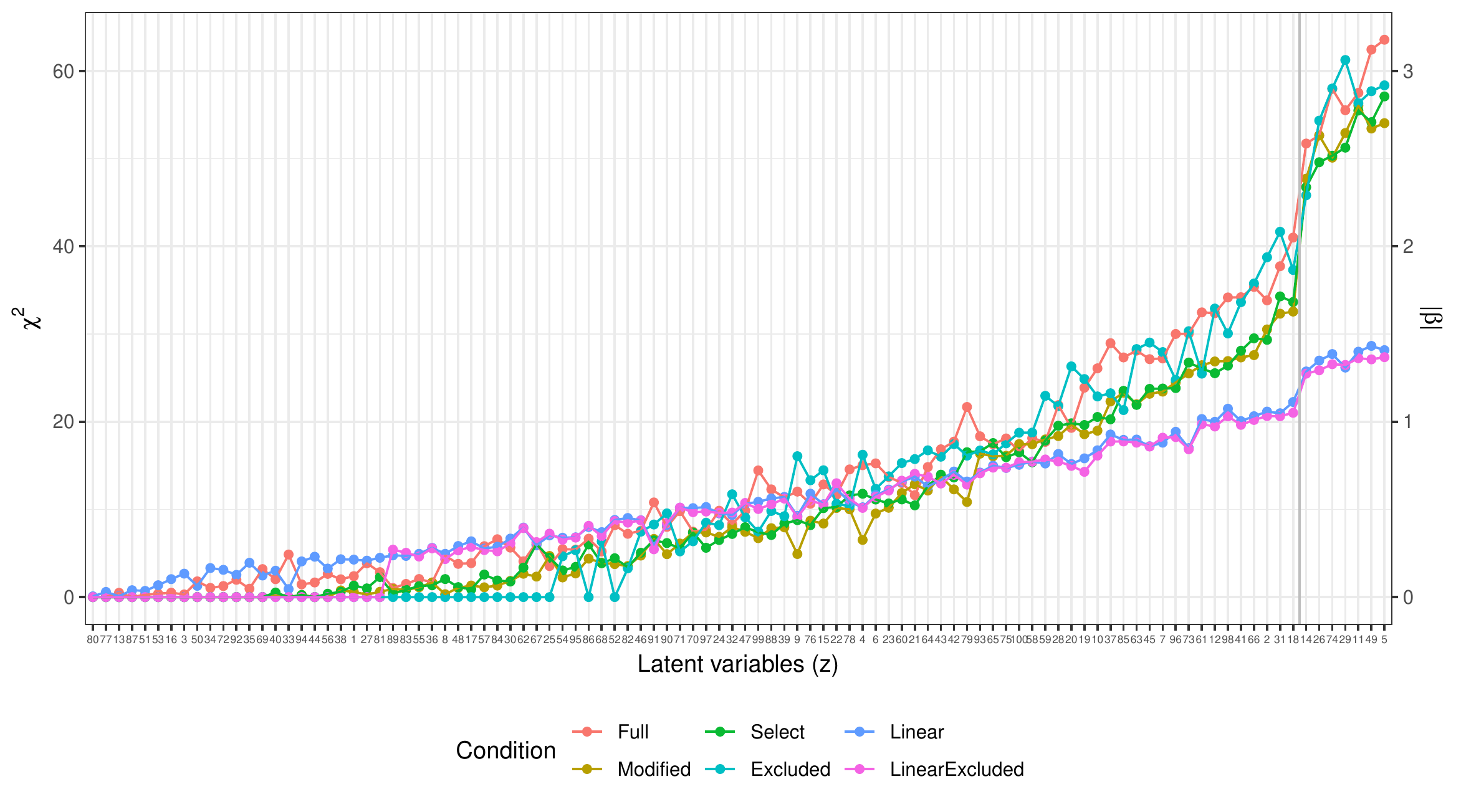}
\caption{\label{gans} Plot of $\chi^2$ values (left scale) for the 100 predictors across the four generalized additive models. For the two linear models (\textsc{Linear} and \textsc{Linear excluded}), estimates of slopes in absolute values ($|\beta|$) are plotted (right scale). The blue vertical line indicates the division between the seven chosen predictors and the rest of the predictor space with a clear drop in estimates between the first seven values ($z_5$, $z_ {11}$, $z_ {49}$, $z_{29}$, $z_ {74}$, $z_{26}$, $z_{14}$) and the rest of the space.}
\end{figure}

To identify the latent variables with the highest correlation with [s] in the output, we extract $\chi^2$ estimates for each term from the generalized additive models and estimates of slopes ($\beta$) from the linear model. Figure \ref{gans} plots those values in a descending order. The plot points to a substantial difference between the highest seven predictors and the rest of the latent space. Seven latent variables are thus identified ($z_5$, $z_ {11}$, $z_ {49}$, $z_{29}$, $z_ {74}$, $z_{26}$, $z_{14}$) as potentially having the largest effect on the presence or absence of [s] in output.  
Figure \ref{gam} plots smooths of the seven predictors ($z_5$, $z_ {11}$, $z_ {49}$, $z_{29}$, $z_ {74}$, $z_{26}$, $z_{14}$) from a non-linear model \textsc{Select}. The smooths show a linear or near-linear relationship between values of the chosen seven variables and  the probability of [s] in the output.

Several methods for finding the features that predict the presence or absence of [s] are thus used. Logistic regression is presented here because it is the simplest and easiest to interpret. In future work, a  combination of techniques is recommended to be used for exploratory purposes in a similar way as proposed in this paper. Below, we conduct two independent generative tests to evaluate whether the proposed technique indeed identifies variables that correspond to presence of [s] in the output.

\subsubsection{Generative test 1}
\label{Generativetest1}

To conduct an independent generative test of whether the chosen values correlate with  [s] in the output data of the Generator network, we set values of the seven identified predictors ($z_5$, $z_ {11}$, $z_ {49}$, $z_{29}$, $z_ {74}$, $z_{26}$, $z_{14}$) to the marginal value of 1 or $-1$ (depending on whether the correlation is positive or negative; see Figure \ref{gam}) and generated 100 outputs. Altogether seven values in the latent space were thus manipulated, which represents only 7\% (7/100) of all latent variables. Of the 100 outputs with manipulated values, 73 outputs included an [s] or [s]-like element, either with the stop closure and vowel or without them.  The rate of outputs that contain [s] is thus significantly higher when the seven values are manipulated to the marginal levels compared to randomly chosen latent space. In the output data without manipulated values, only 271 out of 3800 generated outputs (or 7.13\%) contained an [s]. The difference is significant ($\chi^2(1)=559.0, p<0.00001$).

High proportions of [s] in the output can be achieved with manipulation of single latent variables, but the values need to be highly marginal, i.e.~extend well beyond the training space. Setting the $z_{11}$ value outside the training interval to $-15$, for example, causes the Generator to output [s] in 87 out of 100 generated (87\%) sequences, which is again significantly more than with randomly distributed input latent variables ($\chi^2(1)=792.7, p<0.0001$). When  $z_{11}$ is $-25$, the rate goes up to 96 out of 100, also significantly different from random inputs ($\chi^2(1)=959.8, p<0.0001$).

\begin{figure}
\centering
\includegraphics[width=0.8\textwidth]{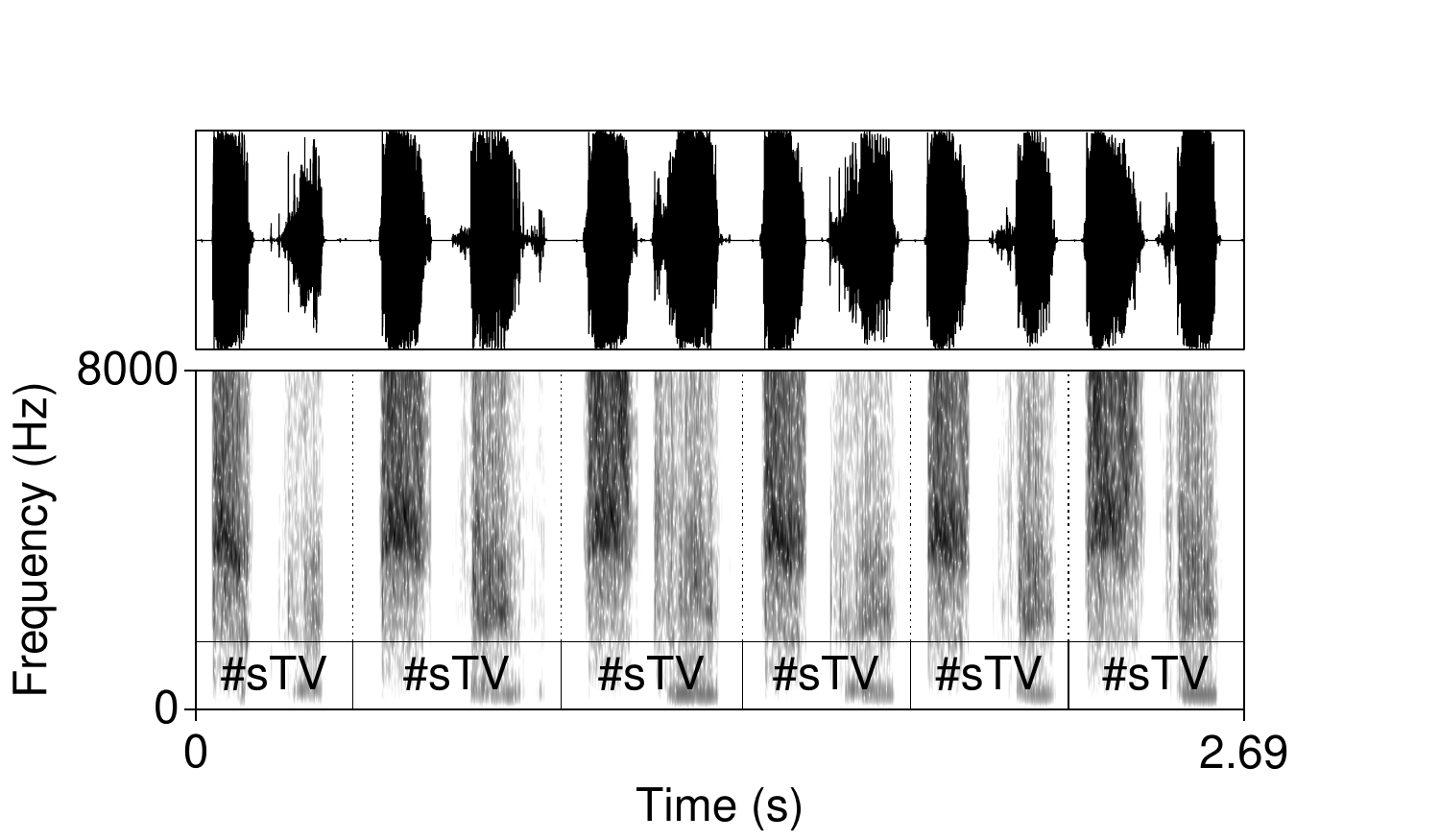}
\caption{\label{sss}Seven waveforms and spectrograms (0-8000 Hz) of outputs from the Generator network with the value of $z_{11}$ set at $-25$. In 96 out of 100 generated samples, the network outputs a sequence containing an [s]. With such a low value of $z_{11}$ (that correlates with amplitude of frication noise), the amplitude of the frication noise reaches the maximum level of 1 in all 100 generated outputs.  }
\end{figure}

\subsubsection{Generative test 2}
\label{gen2}

To further confirm that the regression models identify the variables involved with the presence of [s] in generated outputs, another generative experiment was conducted. In addition to manipulating the seven identified variables, we test the effect of other variables in the latent space on the presence of [s] in the input. If the regression estimates provide reliable information, the variables with higher estimates should have more of an effect on the presence of [s] in the output and vice versa.  Testing the entire latent space would be too expensive, which is why we limit our tests to 25 variables with highest estimates from the regression models (which includes the seven chosen variables) and six additional variables with descending regression estimates.  Altogether 31/100 variables or 31\% of the latent variables are thus analyzed. The  variables were chosen in the following way: first, we manipulate values of the first 25 variables with the highest estimates based on regression models in Figure \ref{gans}  (7 chosen variables plus additional 18 variables for a total of 25). Because we want to test the effects of the latent variable as evenly as possible and also  to test the effects of variables with the lowest regression estimates, we picked 6 additional variables that are distanced from the 25th variable in increments of 5 after the  variable with the 25th highest estimate (random choice of variables might miss the variables with lowest estimates).

To perform the generative test of the correlation between the latent space and the proportion of  [s] in the output, we  set each of the 31  variables at a time to a marginal level well beyond the training interval (to $\pm$4.5), while keeping the rest of the latent space randomly sampled, but constant.  
In other words, all variables are sampled randomly and held constant across all samples, with the exception of the variable in question at a time that is set to $\pm$4.5.  The $\pm$4.5 value was chosen based on manual inspection of generated samples: as is clear from Figure \ref{maxint}, changes in amplitude of [s] become increasingly smaller when variables have a value greater than $\pm$3.5.  For effects of values beyond 4.5, see Figure \ref{sss}. 

\begin{figure}
\centering
\includegraphics[width=0.5\textwidth]{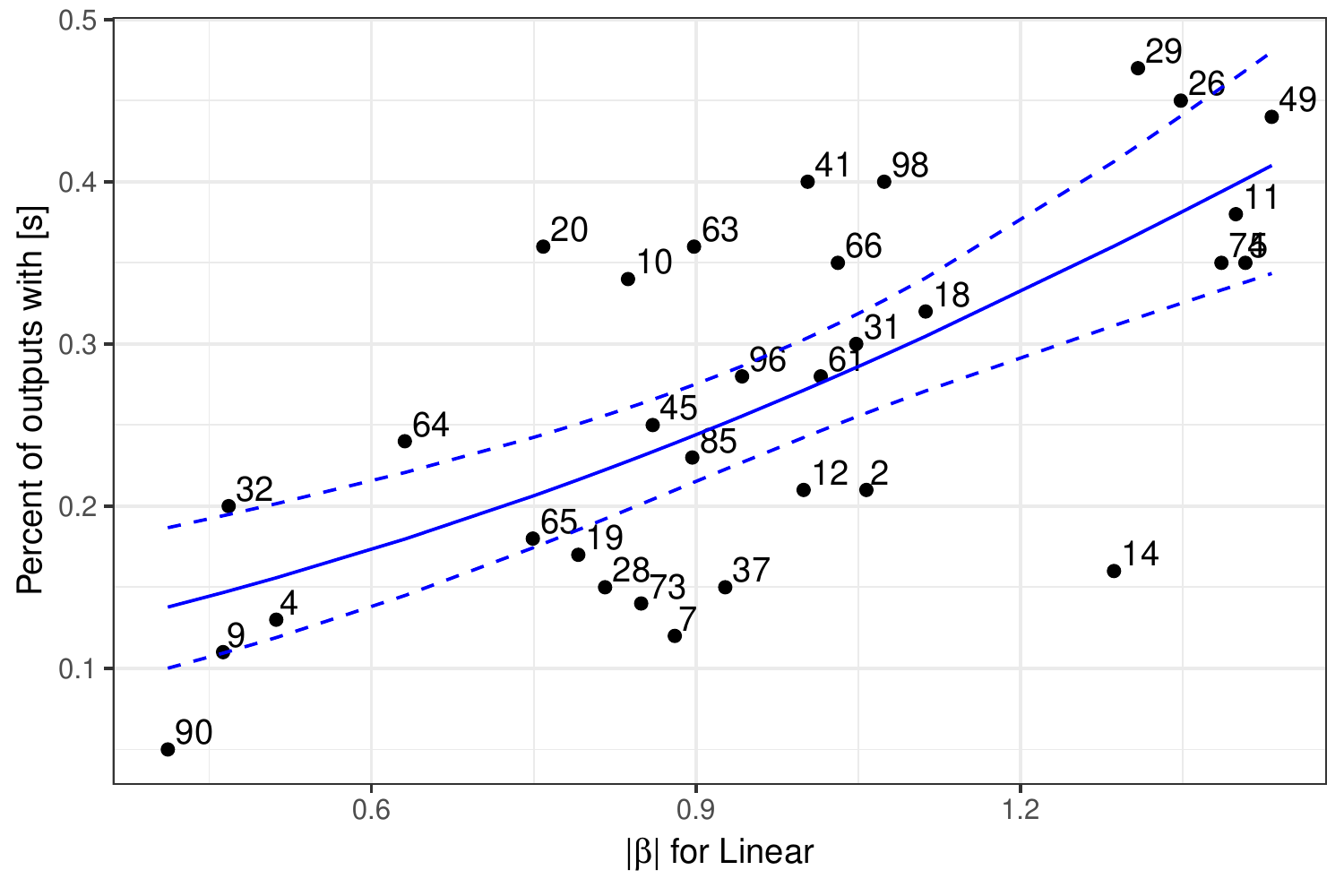}
\caption{\label{zvariables} Plot of absolute values of estimates from the Linear model for 31 analyzed latent variables $z$ (numbered on the plot) and the percent of outputs that contain an [s] based on 100 generated samples. The blue solid line represents predicted values based on the beta regression model with estimates of the Linear model as the predictor; the dashed lines represent 95\% confidence intervals.}
\end{figure}

One hundred outputs are generated for each of the 31 manipulated latent variables. Altogether $31\times100$ (3,100) outputs were thus analyzed and annotated for the presence or absence of [s] in the output. 
For example, when the effect of latent variable $z_{11}$ on the proportion of [s] in the output is tested,   we set its value to $-$4.5 while keeping other variables random.  100 samples are generated in which the other 99 variables are randomly distributed with the exception of the $z_{11}$ variable (which is set at the marginal level). Samples are annotated for the presence or absence of [s] and the proportion of [s] in the output is calculated  from the number of samples with [s] divided by the number of all samples.  The same procedure is applied to the other 30 variables examined. To control for the unwanted effects of the latent space on the output, all 99 other variables with the exception of the one manipulated are kept constant across all 31 samples. The 31 data points of this proportion are thus the dependent variable in regression models (Figure \ref{zvariables}) that test the correlation between the identified variables and [s] in the output.

A beta regression model with the proportion of [s] as the dependent variable and with estimates of the Linear  model as the independent variable suggests that there exists a significant linear correlation between the estimates of the regression models and the actual proportion of generated outputs with [s]: $\beta=1.44, z=5.07, p<0.0001$ (for details on model selection, see Supplementary material Section \ref{supgen2}). In other words, the technique  for identifying latent variables that correlate with the presence of [s] in the output based on regression models (in Figure \ref{gans}) successfully identifies such variables. This is confirmed independently: the proportion of generated outputs containing an [s]   correlates significantly with its estimates from the regression models.

\subsubsection{Interpretation}

The regression models in Figure \ref{gans} identify those $z$-variables in the latent space that have the largest effect on the presence of [s] in the output. Figure \ref{zvariables} confirms that the generator outputs significantly higher proportions of [s] for some variables, while other variables have no effect on the presence of [s]. In other words, variables with lower regression estimates do not affect the proportion of [s] in the output. The proportion of [s] in the output when variables such as $z_{90}$, $z_{9}$,  $z_{4}$,  $z_{7}$ are manipulated is very close to the 7.13\% of [s] in the output when all $z$-variables in the latent space are random.  It thus appears that the Generator uses portions of the latent space to encode the presence of [s] in the output.

Some latent variables cause a high proportion of [s] in the output despite the regression model estimating their contribution lower than the seven identified latent variables (Figure \ref{zvariables}) and vice versa. Outputs for variable $z_{14}$ contain frication noise that falls between [s] and [s]-like aspiration, which were difficult to classify (also, the target for [s]-like outputs in this variable is closer to 2.5). The two variables with the highest proportion of [s] in the output that are estimated substantially lower than the seven variables are $z_{41}$ and $z_{98}$. There is a clear explanation for the discrepancy of the regression estimates and the rates of [s]-outputs for such variables. While outputs at the marginal values of of the two variables (at $\pm4.5$) do indeed contain a high proportion of [s]-outputs, the frication noise ceases during the $(-1, 1)$ interval on which the model is trained. Because the regression model only sees the training interval $(-1, 1)$ (annotations fed to the regression models are performed on this interval) and does not access outputs with variables outside of this interval, the estimates are consequently lower than the outputs at the marginal levels for these variables.  There are only a handful of such variables, and since we are primarily interested in those variables that correspond to [s] both within the training interval and outside of it, we focus our analysis below on  the seven variables identified in Section \ref{inter}. The problem with variables in which [s] outputs are present predominantly outside of the training interval is the possibility that the [s]-output in these types of cases is secondary/conditioned on some other distribution, because it was likely not encoded in the training stage.

While there is a consistent drop in estimates of the regression models after the seven identified variables  (Figure \ref{gans}) and while several independent generation tests confirm that the seven variables have the strongest effect on  the presence of [s] in the output, the cutoff point between the seven variables and the rest of the latent space is still somewhat arbitrary. It is likely that other latent variables directly or indirectly influence the presence of [s] as well: the learning at this point is not yet categorical and several dependencies not discovered here likely affect the results. Nevertheless, further explorations of the latent space suggest the variables identified with the logistic regression (and other) models (Figure \ref{gans}) are indeed the main variables involved with the presence or absence of [s] in the output.

Additionally, if at the value of $z$ that so substantially exceeds the training interval ($\pm4.5$) the latent variable does not influence the outcomes substantially and only marginally increases the proportion of [s]-outputs, as is the case for the majority of the latent variables outside of the seven chosen ones, it is likely that its correlation with [s] in the output is secondary and that the variable does not contribute crucially to the presence of [s].

\subsection{\label{inter}Interpolation and phonetic features}

Fitting the annotated data and corresponding latent variables from the Generator network to generalized additive and linear logistic regression models identifies values in the latent space that correspond to the presence of [s] in the output.   As will be shown below, this is not where exploration of the Generator's internal representations should end. We explore whether the mapping between the uniformly distributed input $z$-variables and the Generator's output signal that resembles speech can be associated with specific phonetic  features in that output. The crucial step in this direction is to explore values of the latent space and their phonetic correlates in the output beyond the training interval, i.e.~beyond $(-1,1)$. We observe that the Generator network, while being trained on latent space limited to the interval $(-1, 1)$, learns representations that extend this interval. Even if the input latent variables ($z$) exceed the training interval, the Generator network outputs samples that closely resemble human speech. Furthermore, the dependencies learned during training  extend outside of the $(-1, 1)$ interval. As is argued in Section \ref{pvlv}, exploring phonetic properties at these marginal values  has the potential to reveal the actual underlying function of each latent variable.

\begin{figure}
\centering
\includegraphics[width=0.9\textwidth]{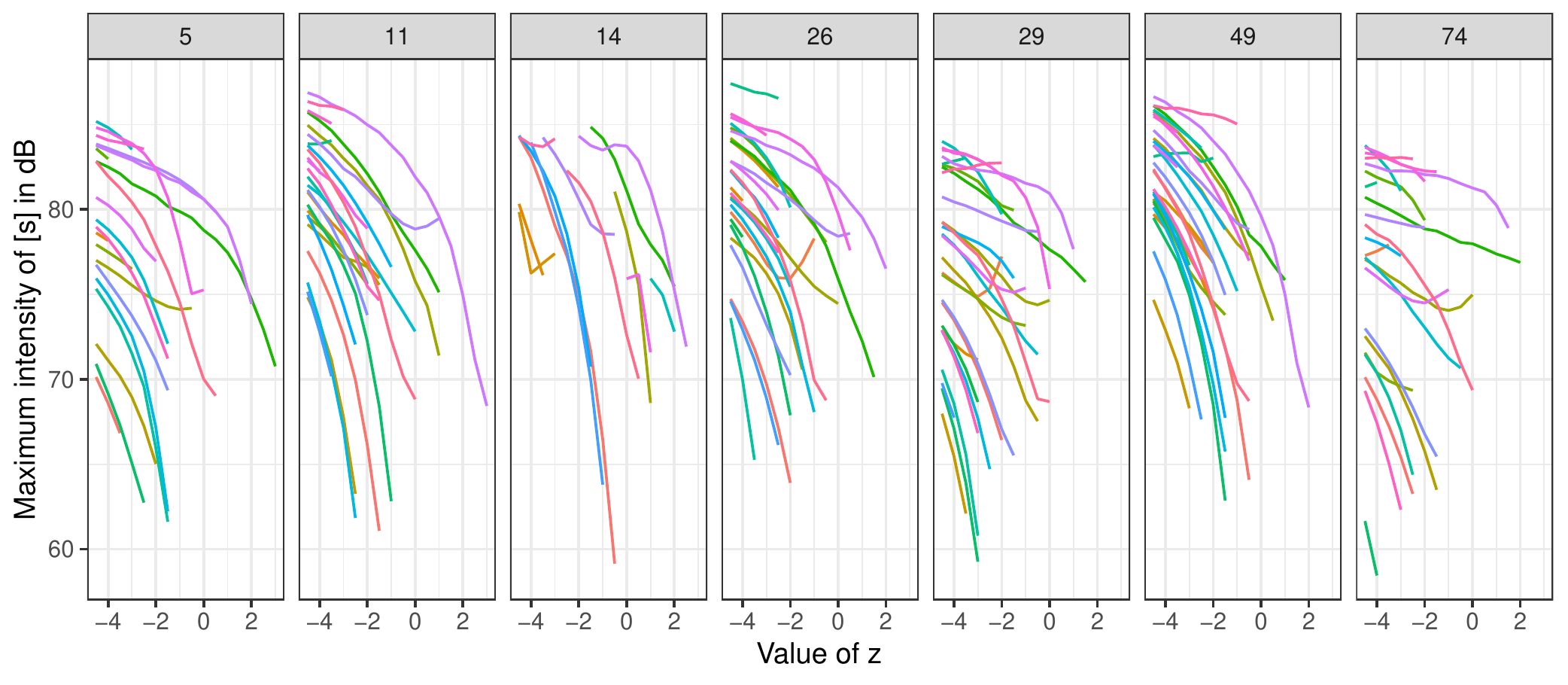}
\caption{\label{maxint} Plots of maximum intensity (in dB) of the fricative part in \#sTV sequences when values of the seven $z$-variables are interpolated from the marginal values $\pm$4.5 in 0.5 increments. Each set of generated samples with the randomly sampled latent variables held constant is colored with the same color across the seven $z$-variables. Values of $z_5$, $z_{14}$,  and $z_{26}$ are inverted for clarity purposes.}
\end{figure}

To explore phonetic correlates of the seven latent variables, we set each of the seven variables separately to the marginal value $-4.5$ and interpolate to its opposite marginal value 4.5 in 0.5 increments, while keeping randomly-sampled values of the other 99 latent variables $z$ constant. Again, the $\pm$4.5 value was chosen based on manual inspection of generated samples: amplitude of [s] ceases to change substantially past values around  $\pm$3.5 (Figure \ref{maxint}).  Seven sets of generated samples are thus created, one for each of the seven $z$ values: $z_5, z_{11}, z_{14},  z_{26},  z_{29},  z_{49}$, and  $z_{74}$ (with the other 99 $z$-values randomly sampled, but kept constant for all seven manipulated variables). Each set contains a subset of 19 generated outputs that correspond to the interpolated variables from $-4.5$ to 4.5 in 0.5 increments. Twenty-nine such sets that contained an [s] in at least one set are extracted for analysis (sets that lack an [s] were not analyzed).

A clear pattern emerges in the generated data: the latent variables identified as corresponding to the presence of [s] via regression (Figure \ref{gans}) have direct phonetic correlates and cause changes in amplitude and the presence/absence of frication noise of [s] when each of the seven values in the latent space are manipulated to the chosen values, including values that exceed the training interval. In other words, by manipulating the identified latent variables, we control the presence/absence of [s] in the output as well as the  amplitude of its frication noise.

\begin{figure}
\centering
\includegraphics[width=0.49\textwidth]{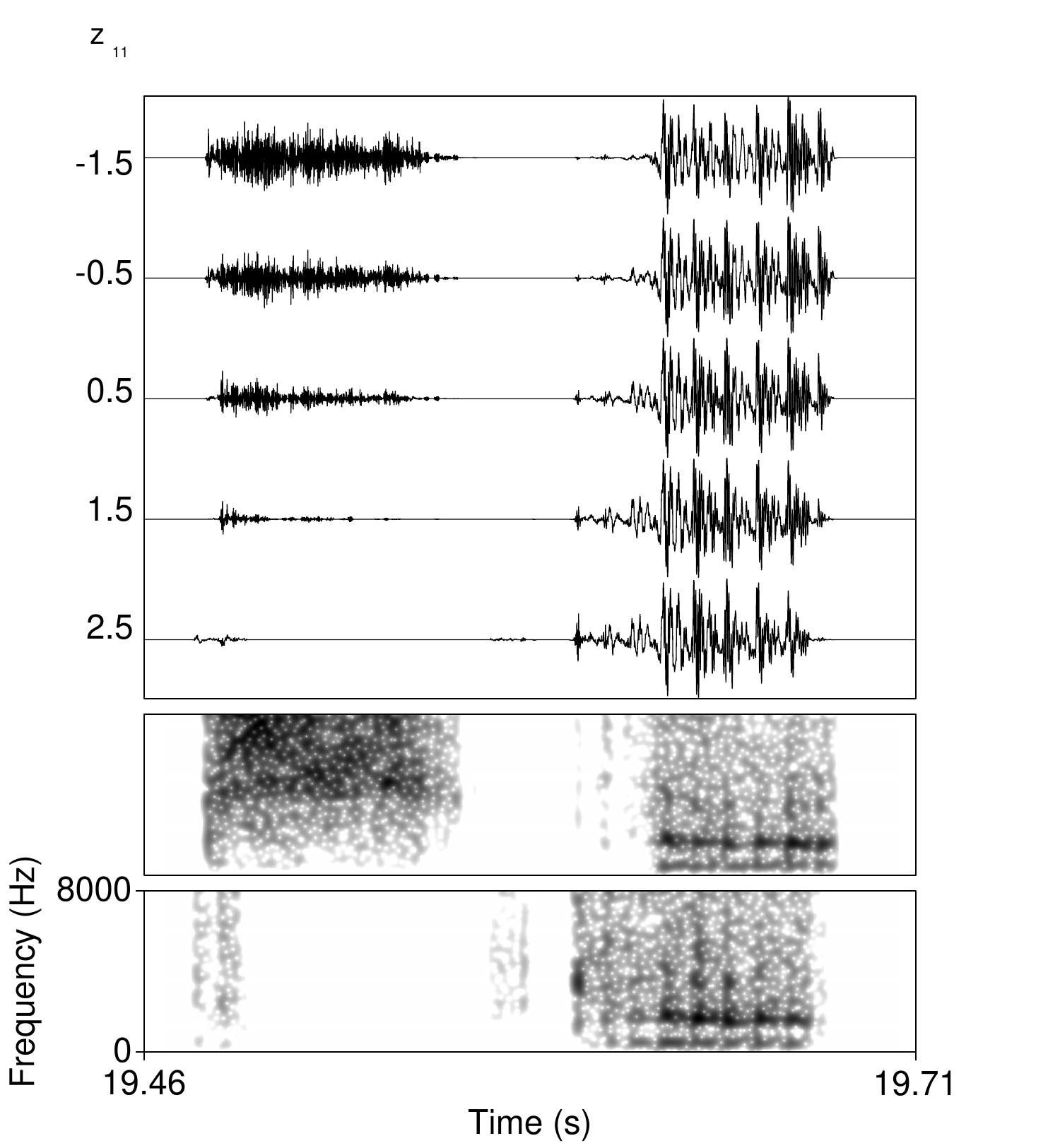}\includegraphics[width=0.49\textwidth]{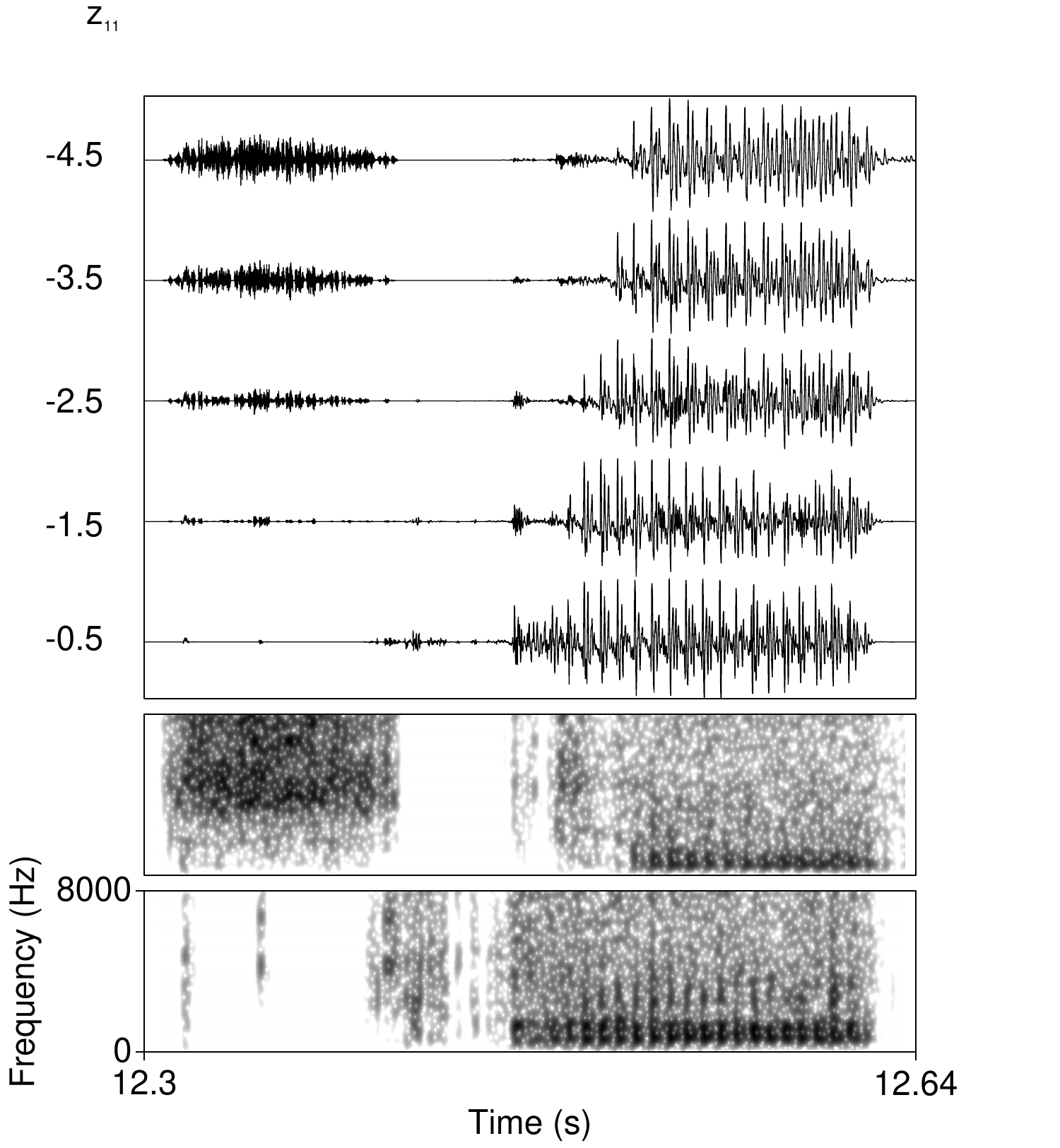}
\caption{\label{interpolation2}Waveforms and two spectrograms (both $0-8,000$ Hz) of generated data with $z_{11}$ variable manipulated and interpolated. The values on the left of waveforms indicate the value of $z_{11}$. The two spectrograms represent the highest and the lowest value of $z_{11}$. A clear attenuation of the frication noise is visible until complete disappearance.}
\end{figure}

Figure \ref{interpolation2} illustrates this effect. Frication noise of [s]  gradually decreases by increasing the value of $z_{11}$ until it completely disappears from the output. The exact value of $z_{11}$ for which the [s] disappears differs across examples and likely interacts with other features.  It is possible that frication noise in the training has a higher amplitude in some conditions, which is why such cases require a higher magnitude of manipulation of $z_{11}$. The figure also shows that as the frication noise of [s] disappears, aspiration of a stop in what appear to be \#TV sequences   starts surfacing and replaces the frication noise of [s]. Occasionally, frication noise of [s] gradually transforms into aspiration noise. The exact transformation is likely dependent on the 99 other $z$-variables held constant and their underlying phonetic effects. Regardless of these underlying phonetic effects, manipulating the chosen variables has a clear effect of causing [s] to appear in the output and controlling its amplitude.

To test the significance of the effects of the seven identified features on the presence of [s] and the amplitude of its frication noise, the 29 generated sets of 19 outputs (with $z$-value from $-4.5$ to $4.5$) for each of the seven variables were analyzed. The outputs were manually annotated for [s] and the following vowel.  Outputs gradually change from \#sTV to \#TV. Only sequences containing an [s] were analyzed; as soon as [s] stops in the output, annotations were stopped and the outputs were not further analyzed. Altogether 161 trajectories were thus annotated; the total number of data points measured is  1,088 because each trajectory contains a number of measurements of the interpolated values of $z$. For each datapoint, maximum intensity of the fricative and maximum intensity of the vowel were extracted in Praat \citep{boersma15} with a 13.3 ms window length (with parabolic interpolation).\footnote{The script used for this task was provided by \cite{lennes03}.}  Figure \ref{maxint} illustrates how manipulating the values of $z$ of the chosen variables from the marginal value $-4.5$ decreases frication noise in the output until [s] is completely absent. 

\begin{figure}
\centering
\includegraphics[width=0.8\textwidth]{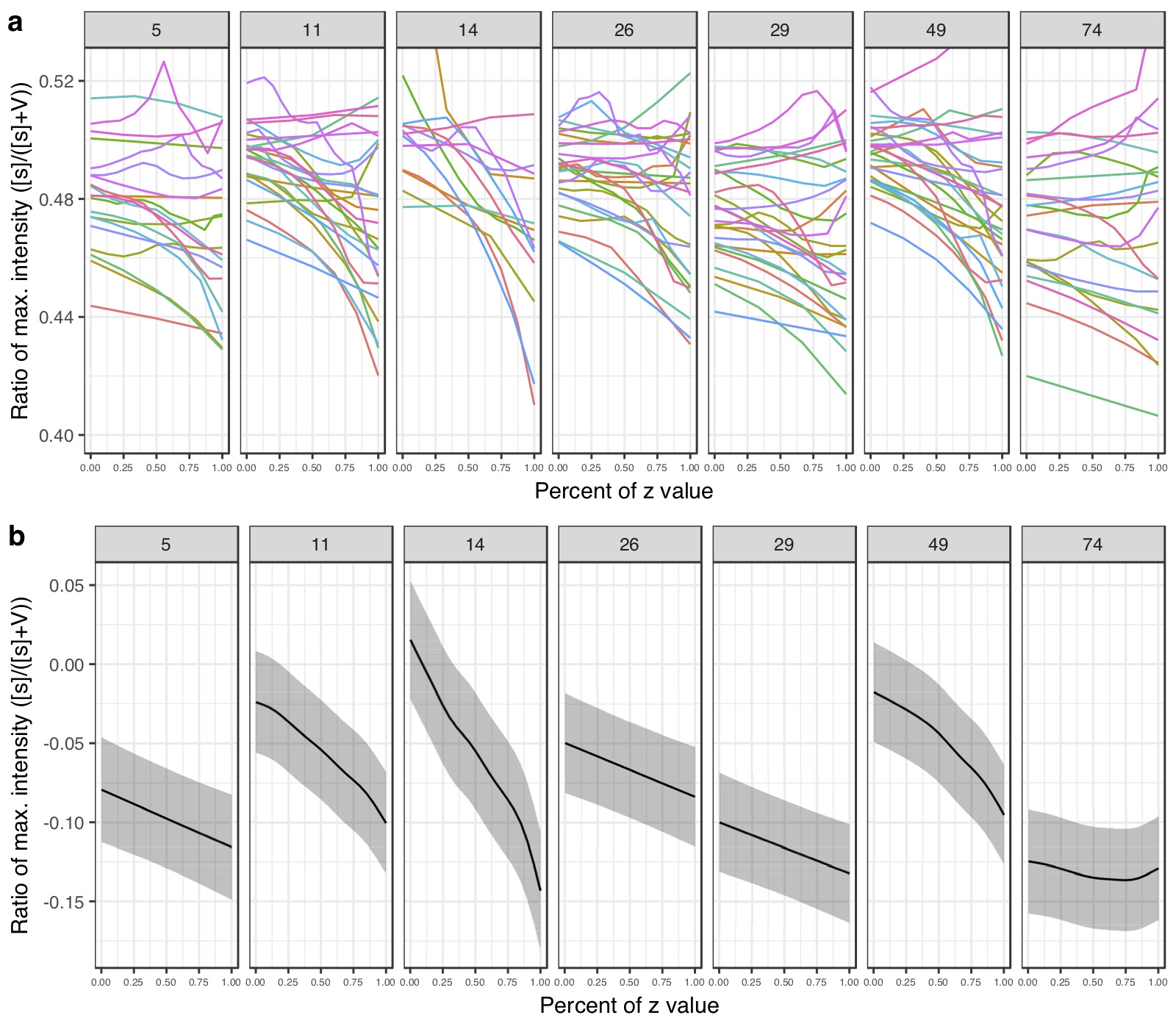}
\caption{\label{ratio12} \textbf{(a)} Plots of ratios of maximum intensity between the frication of [s] and phonation of the vowel in \#sTV sequences across the seven variables. The interpolated values are normalized where 0 represents the most marginal value of $z$ with [s] in the output and 1 represents the value of $z$ right before [s] ceases from the output. Four marginal values are left out from the plot (but are included in the models). Each set of generated samples with the randomly sampled latent variables held constant is colored with the same color across the seven $z$-variables. \textbf{(b)} Predicted values with 95\% CIs of the ratio based on beta regression generalized additive model (Table \ref{zValueGam}) across the several variables with normalized values. Values of $z_5$, $z_{14}$,  and $z_{26}$ are inverted for clarity purposes.}
\end{figure}

To test whether the decreased frication noise is not part of a general effect of decreased amplitude, we perform significance tests on the ratio of maximum intensity between the frication noise of [s] and the following vowel in the \#sTV sequences. Figure \ref{ratio12} plots the ratio of maximum intensity of the fricative divided by the sum of two maximum intensities: of the fricative ([s]) and of the vowel (V). The manipulated $z$-values are additionally normalized to the interval [0,1], where 0 represents the most marginal value with [s] (usually $\pm4.5$; referred to as \textsc{strong} henceforth) and 1 represents the last value before [s] disappears (\textsc{weak}). Note that the point at which [s] is not present in the output anymore, but the vowel still surfaces (which would yield the ratio at 0) is not included in the model. 

The data were fit to a beta regression generalized additive mixed model (in the \emph{mgcv} package; \citealt{mgcv}) with the ratio as the dependent variable, the seven chosen variables as the parametric term, thin-plate smooths for each variable and random smooths (with first order of penalty; \citealt{baayen16,soskuthy17}) for (i) trajectory and for  (ii) value of other variables in the latent space of the Generator network. Figure \ref{ratio12} plots the normalized trajectories of the ratio and predicted values based on the generalized additive model. All smooths (except for $z_{74}$) are significantly different from 0 (all coefficients in Table \ref{zValueGam}) and the plots show a clear negative trajectory. In other words, maximum intensity of [s] is increasingly attenuated compared to the intensity of the vowel as $z$ approaches the opposite value from the one identified as predicting the presence of [s] until it completely disappears from the output.

The seven variables thus strongly correspond to the presence or absence of [s] in the output; by manipulating the chosen variables  to the identified values we can attenuate frication noise of [s] and cause its presence or complete disappearance in the generated data. Again, the discovery of these features is possible because we extend the initial training interval and test predictions on marginal values. In Section \ref{pvlv}, we  analyze further phonetic correlates of each of the seven variables.

\subsection{\label{pvlv}Phonetic values of latent variables}

Interpolation of latent variables reveals that the presence of [s] is not controlled by a single latent variable, but by at least seven of them. Additionally,  there appears to be no categorical cut-off point in the magnitude of the effect between the variables, only a steep drop of regression estimates (Figure \ref{gans}) and a  decline of outputs with [s] in generated data (Figure \ref{zvariables}). This suggests that the learning at this stage is gradient and probabilistic rather than fully categorical. 

The different latent variables that correspond to the presence of [s], however, are not phonetically vacuous: individually, they have distinct phonetic correspondences.
The generated samples reveal that the variables' secondary  effect (besides outputting [s] and controlling its intensity) are likely  spectral properties of the frication noise. The seven variables are thus  similar in the sense that manipulation of their values results in the presence of [s] by controlling its frication noise. They crucially differ, however, in the effects on the spectral properties of the outputs. 

To test this prediction, spectral properties of the output fricatives are analyzed.  The same 29 sets of generated samples are used in the analysis; one $z$-value is manipulated in  each set while other variables are sampled randomly and held constant. The marginal values of the variables were chosen for this test: the values with the strongest presence of [s] (which in most cases is $\pm$4.5; henceforth \textsc{strong}) and the value before which [s] ceases from the output (henceforth \textsc{weak}). Center of gravity (COG), kurtosis, and skew of the frication noise were analyzed with and extracted with a script from \citet{rentz17} in Praat \citep{boersma15}.  Period of frication is sliced into 10\% intervals. The data includes 161 trajectories (from the 29 generated sets) and $161\times10=1,610$ unique data points. COG, kurtosis, and skew based on power spectra are measured in each of these 1,610 intervals with 750--8,000 Hz Hann band pass filter (100 Hz smoothing).  Results were fit to six generalized additive mixed models with COG, kurtosis, and skew as the dependent variables (3 for each of the levels \textsc{strong} and \textsc{weak}). The parametric terms included the seven latent variables $z$. The smoothing terms included smooths for  latent variable $z_{11}$ and difference smooths for the other six variables $z$. The model also includes random smooths for each fricative (from 10 to 100\% with 10 knots) and for each of the 29 generated sets with equal random values of other 99 $z$-variables (with 7 knots; random smooths are fitted with first order of penalty, see \citealt{baayen16,soskuthy17}). The models were fit with correction for autocorrelation with $\rho$-values ranging from 0.15 to 0.7.

\begin{figure}
\centering
\includegraphics[width=0.8\textwidth]{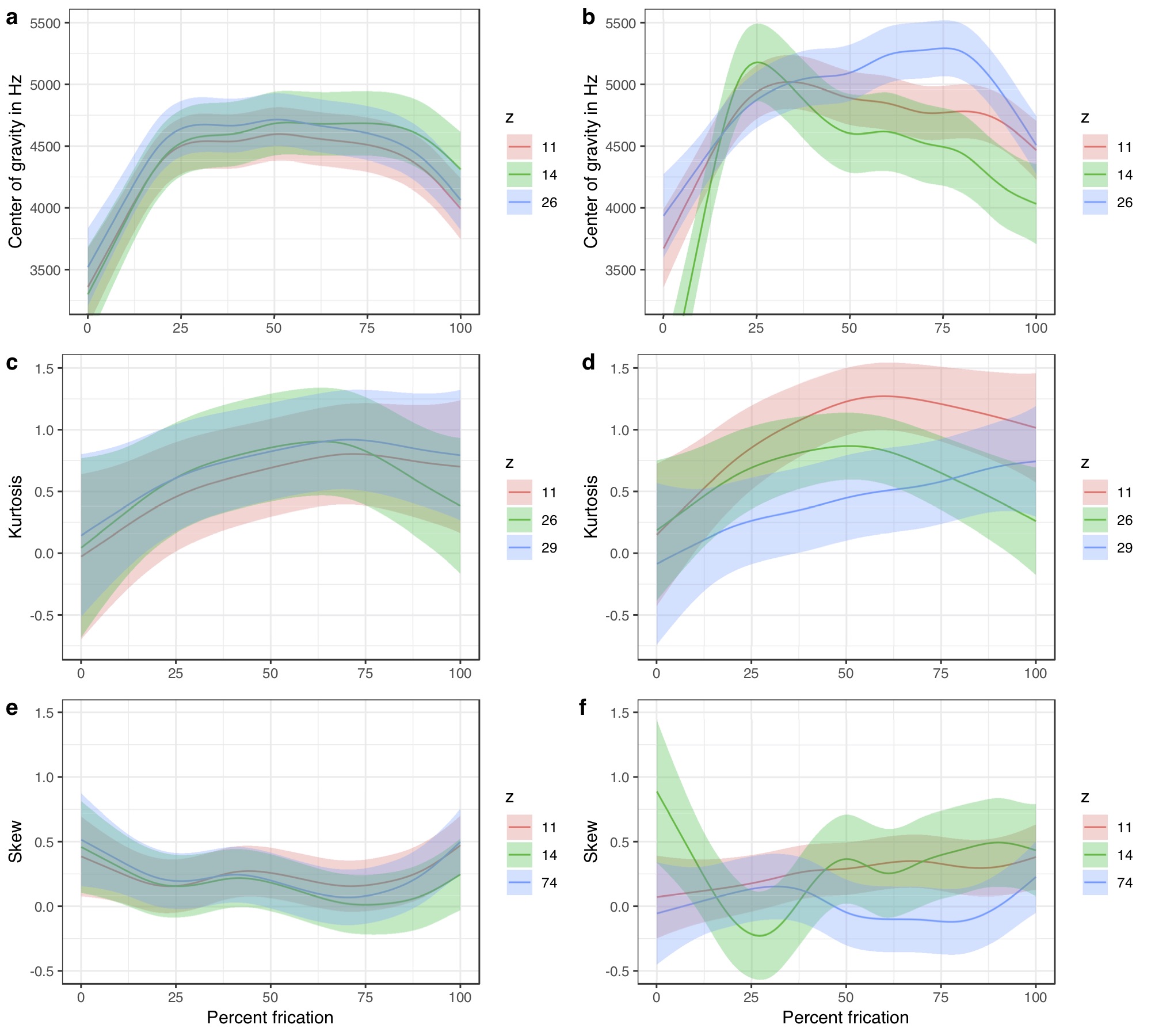}
\caption{\label{kurt}A subset of predicted values of COG, kurtosis, and skew with 95\% CIs in two conditions: \textsc{weak} with $z$-variables at the value before [s] ceases from the output (right column) and \textsc{strong} with the most marginal value with [s]-output ($\pm4.5$ in most cases). Predicted values are based on generalized additive models in  Tables \ref{zValueCOGGamARsum}, \ref{zValueKURTGamARsum}, \ref{zValueSKEWGamARsum}, \ref{zValueKURTGamsum1}, \ref{zValueKURTGamARsum1}, and \ref{zValueSKEWGamARsum1}. The plots show a clear differentiation from no significant differences in COG, kurtosis, and skew, to clear significant overall differences and trajectory differences as the $z$-values move from \textsc{weak} toward the marginal (\textsc{strong}) values. Difference smooths for the presented variables are in Figure \ref{differences}.}
\end{figure}

Spectral properties of the generated fricatives are generally not significantly different at the value of $z$ right before [s] disappears from the outputs (\textsc{weak}; left column in Figure \ref{kurt}). As values of $z$ increase toward the marginal levels (in most cases, $\pm$4.5), however, clear differentiation in spectral properties emerge between some of the seven $z$-variables (\textsc{strong}; right column in Figure \ref{kurt}). The trajectory for center of gravity, for example, significantly differs between $z_{11}$ and most of the other six variables. Overall kurtosis is significantly different when $z_{11}$ is manipulated, compared to, for example, $z_{26}$ and $z_{29}$. Similarly, while $z_{74}$ does not significantly attenuate amplitude of [s], it significantly differs in  skew trajectory of [s]. The main function of $z_{74}$ is thus likely in its control of spectral properties of frication of [s] (e.g.~skew). For all coefficients and significant  and non-significant relationship of the six models, see Tables \ref{zValueCOGGamARsum}, \ref{zValueKURTGamARsum}, \ref{zValueSKEWGamARsum}, \ref{zValueKURTGamsum1}, \ref{zValueKURTGamARsum1}, and \ref{zValueSKEWGamARsum1}.

In sum, manipulating the latent variables that correspond to [s] in the output not only attenuates frication noise (when vocalic amplitude is controlled for)  and causes [s] to surface or disappear from the output, but the different $z$-variables likely correspond to different phonetic features of the frication noise. At the level before the frication noise ceases from the output, there are no differences in spectral moments between the latent variables. By setting the values to the marginal levels well beyond the training interval, however, significant differences emerge both in overall levels as well as in trajectories of COG, kurtosis, and skew. It is thus likely that the variables collectively control the presence or absence of [s], but that individually, they  control various phonetic features --- spectral properties of the frication noise.

\section{Discussion}
\label{discussion}

The Generator network trained after 12,255 steps learns to generate outputs that closely resemble human speech in the training data. 
The results of the experiment in Section \ref{m12255} suggest that the generated outputs from the Generator network replicate the conditional distribution of VOT duration in the training data. The Generator network thus not only learns to output signal that resembles human speech from noise (input variables sampled from a uniform distribution), but also learns to output shorter VOT durations when [s] is present in the signal. While this distribution is phonologically local, it is non-local in phonetic terms as a period of closure necessarily intervenes between [s] and VOT. 
It is likely, however, that minor local dependencies (such as burst or vowel duration) also contribute to this distribution. 
While it is currently not possible to disambiguate between the local and non-local effects, it is desirable for a model to capture all possible dependencies, as speech production and perception often employ several cues as well.

\subsection{\label{parallelshuman}Parallels in human behavior}

Several similarities emerge between the training of the Generative Adversarial networks and L1 acquisition. The training data in the GAN model is of course not completely naturalistic (even though the inputs are raw audio recordings of speech data): the network is trained on only a subset of sound sequences that a language learning infant is exposed to. The purpose of these comparisons is not to suggest the GAN model learns the data in exactly the same manner as human infants, but to suggest that clear similarities exist in behavior between the proposed model and human behavior in speech acquisition. Such comparisons have both the potential to inform computational models of human cognition and conversely, shed light on the question of how neural networks learn the data.

While the generated outputs contain evidence that the  network learns the conditional distribution of VOT duration, a proportion of outputs violates this distribution. In fact, in approximately 12.2\% of the \#sTV sequences,  the Generator outputs VOT durations  that are longer than any VOT duration in the \#sTV condition in the training data. This suggests that the model learns the conditional distribution, but that the learning is imperfect and the Generator occasionally violates the distribution. Crucially, these outputs that violate the training  data closely resemble human behavior in L1  acquisition. Infants acquiring VOT in English undergo a period in which they produce VOT durations substantially longer compared to the adult input, not only categorically in all stops \citep{macken80,catts83,lowenstein08}, but also in the position after the sibilant [s]. \cite{mcleod96} studied acquisition of \#sTV and \#TV sequences  in 2;0 to 2;11 year old children. Unlike the Generator network, children often simplify the initial clusters  from \#sTV  to a single stop \#TV. What is  parallel to the outputs of the Generator, however, is that the VOT duration of the simplified stop is overall significantly shorter in underlying  \#sTV sequences, but there exists a substantial period of variation and occasionally the language-acquiring children output long-lag VOT durations there (\citealt{mcleod96}, for similar results in language-delayed children, see \citealt{bond81}). \cite{bond80} present a similar study, but include older children that do not simplify the \#sT cluster. This group behaves exactly parallel to the Generator's network: the overall duration of VOT in the \#sTV sequences is shorter compared to the \#TV sequences, but the longest duration of any VOT is attested once in the  \#sTV, not in the \#TV condition \citep{bond80}. The children thus learn both to articulate the full \#sT cluster and to output a shorter VOT durations in the cluster condition. Occasionally, however, they output a long-lag VOT in the  \#sTV condition that violates the allophonic distribution and is longer than any VOT in the \#TV condition.

Further parallels exist between the Generator's behavior and L2 acquisition and speech errors. Studies on L2 acquisition of VOT durations in \#sTV and \#TV sequences suggest that learners start with a smaller distinction between the two groups and acquire the non-aspiration rule after [s] only with more exposure \citep{haraguchi03}. A smaller initial difference between the two conditions in L2 acquisition, for example, increases from Japanese learners of English with little exposure when compared with learners with more exposure  \citep{haraguchi03}. Saudi Arabic L2 learners of English produce substantially longer VOT durations in \#sTV sequences compared to the native inputs \citep{alanazi18}, which resembles imperfect learning in the Generator's network.  Speech errors also provide a parallel to the described behavior of the Generator network. German has a similar process of aspiration distribution as English. In an experiment of elicited speech errors, German speakers produced aspirated stops with longer VOT durations in erroneous sequences with inserted sibilant in 34\% of cases  \citep{pouplier14}. This suggests that the allophonic rule fails to apply in the speech errors, which is parallel to the Generator network outputting a long VOT in the \#sTV condition that violate the training data distributions.

Finally, the Generator network violating the VOT distribution resembles the behavior of patients with speech impairments.  \cite{buchwald12} analyzed VOT durations of two patients with apraxia of speech that present cluster production errors, i.e.~clusters of the structure \#sTV are simplified to \#TV. One patient outputs long VOT durations in the \#sTV condition (after the cluster is simplified). VOT durations in the \#sTV clusters in this patient correspond to VOT durations of singleton stops (\#TV). The other patient also simplifies the cluster, but outputs shorter VOT durations in the \#sTV condition, maintaining the underlying distribution. It is hypothesized that the first patient (with long VOT durations in the \#sTV condition) shows signs of impairment that operates on the phonological level: because phonological computation is impaired, the patient fails to output shorter VOT durations in the \#sTV condition. In other words, there are no motor planning mechanisms that would prevent the patient from producing shorter VOT durations in the \#sTV condition, which is why the error is assumed to operate on the phonological level --- a phonological rule fails to apply, which results in long VOT in the \#sTV condition.   The second patient, on the other hand, is hypothesized to shows traces of phonetic execution impairment, while the phonological computation (short VOT in the \#sTV condition) is intact. The outputs of the Generator network that violate the training data are parallel to the behavior of the patient with assumed phonological impairment: in 12.2\% of cases, the network outputs long VOT duration in the \#sTV condition that is longer than any VOT duration in the same condition in the training data. Since the network lacks any articulatory component (see also discussion below), motor planning factors cannot explain the Generator's violations of the distributions in the training data.

As indicated by examples in Figures \ref{lepS} and \ref{pta}, the network also generates segmentally innovative outputs for which no evidence was available in the training data.  
A subset of the innovative  outputs, such as \#sV and \#TTV sequences, are consistent with linguistic behavior in humans. The Generator's innovative outputs thus closely resemble one of the main properties of human phonology:  productivity. Human subjects are able to evaluate and produce nonce-words even if a string of phonemes violates language-specific phonotactics, as long as the basic universal phonotactic requirements that treat phones as atomic units are satisfied (for an overview of phonotactic judgments, see \citealt{ernestus11} and literature therein). Deleting or inserting segments are also common patterns in both L1 acquisition \citep{macken81}, loanword phonology \citep{yildiz05},  in children with speech disorders \citep{catts84,barlow01},  as well as in speech errors \citep{aldereteWIRE}. For example, \#sT clusters are often simplified in L1 acquisition \citep{gerlach10}. While the most common outcome is deletion of [s] (which results in the \#TV sequence), deletion of the stop is robustly attested as well in L1 acquisition (resulting in \#sV), both in the general population and in infants with speech disorders \citep{catts84,ohala99,gerlach10,syrika11}. While this deletion likely involves articulatory factors that are lacking in our model, the fact that segmental units can be deleted from the output and recombined in L1 acquisition resembles the deletion in  the Generator's  innovative outputs, such as the  \#sV sequence. 

 These innovative outputs of the Generator's network have potential for contributing to our understanding of the evolution of phonology in language evolution in general (for an overview of the field, see \citealt{gibson12}). The main process that any model of the evolution of phonology needs to explain is the change from ``holistic'' acoustic signals in the proto-language to the ``combinatorial'' principle that operates with discrete units --- phonemes and their combinations \citep{oudeyer01,oudeyer02,oudeyer05,oudeyer06,zuidema09}. The Generator network shows traces of this behavior: in addition to learning to reproduce the input, it learns to recombine segments into novel and unobserved sequences.  The exact details of modeling phonological evolution with Generative Adversarial architecture is, however,  beyond the scope of the present paper.
 
 \subsection{Latent variables as correlates of features}

In Section \ref{internalrep}, we  propose a technique for recovering internal representations of the Generator network.  The first crucial observation is that the dependencies learned in the latent space limited by some interval extend beyond that interval. This allows for an in-depth analysis of phonetic effects of each latent variable in the generated data. Regression models  identify those variables in the latent space that strongly correlate with the presence of [s] in the output. Manipulating values of the identified latent variables, both within the training interval and outside of it, results in significantly higher rates of [s] in the output. By interpolating values of individual latent variables outside of the training interval, we explore the exact phonetic correlates of each latent variable. The results suggest that the Generator network learns to use latent variables to encode imperfect equivalents of phonetic features. Since the features not only correspond to phonetic properties, but to the categorical presence or absence of [s] in the output, the network not only uses latent space to encode what would be an approximate equivalent of phonological representations in the broadest sense --- absence or presence of a segment. 

While the presence of [s] in the output is controlled by multiple latent variables, each of the variables likely has an underlying phonetic function. While there are no significant differences in phonetic correlates of $z$-variables when their value is at the last point before [s] ceases from the output, a clear differentiation emerges when the values are set to the marginal level (Figure \ref{kurt}). The seven variables thus likely have a phonetic function:  controlling various spectral properties of the frication noise.

Features have long been in the center of phonetic and  phonological literature \citep{trubetzkoy39,ch68,clements85,dresher15,shain19}.  Extracting features based on unsupervised learning of pre-segmented phones with neural networks has recently seen success in the autoencoder architecture \citep{rasanen16,eloff19,shain19}. \cite{shain19} train an autoencoder with binary stochastic neurons on pre-segmented speech data and argue that  bits in the code of the autoencoder network imperfectly correspond to phonological features as posited by phonological theory.  As was argued in Section \ref{internalrep},  our model shows traces of imperfect  self-organizing of phonetic features (e.g.~spectral moments) and phonological representations (e.g.~the presence of [s])  in the latent space, while learning allophonic distributions at the same time. Considerable differences between the theoretically assumed features and our results, of course, remain. Latent space encoding in our model resembles entire phonological feature matrices (such as the full presence of [s] in the output) and phonetic features (such as COG or kurtosis), but the relationships are gradient and not categorical. The current model also does not test whether higher order grouping of phonemes in accordance with actual phonological  features such as [$\pm$sonorant] emerge in the training. This task is left for future work. Despite these differences, the fact that we can actively control the presence of [s] and its spectral properties in the generated data with a subset of latent variables suggest that the network learns to encode information in its latent space that resembles phonetic and phonological representations.

On a very speculative level, the latent space of the Generator's network might have a conceptual correlation in featural representation of speech production in human brain, where featural representations are also gradient and involve multiple correlates. \cite{Bashivaneaav9436} argue for the existence of direct correlations between the neural network architecture and vision in human brain. Similarly, \cite{guenther12}, \cite{guenther16}, and \cite{oudeyer05} propose models of simple neural maps that might have direct equivalents in neural computation of speech planning with some actual clinical applications that result from such models.\footnote{\cite{warlaumont16} propose a model of infant babbling that involves spiking neural networks and speech synthesis. While the model does not take any speech as an input, babbling emerges even if the objective for the simulation is maximization of perceptual salience.} Recently, high-density direct cortical surface (electrocorticographic) recordings of the superior temporal gyrus during open brain surgery in \cite{mesgarani14} suggests that recorded brain activity has direct correlates in encoding of phonetic features. Encoding for phonetic and phonological features in the latent space of the Generator's network can speculatively be compared to such brain recordings that serve as the basis for articulatory execution. The correspondences between the brain activity and phonetic and phonological features are multiple and gradual, not categorical, which bear resemblances  to our model. To be sure, this comparison can only be indirect and speculative at this point. 

\subsection{Future directions}

Among the objections against modeling phonological learning with Generative Adversarial Networks might be that the model is too powerful and that it overgenerates. First, it has been shown in numerous examples that phonology, while being computationally limited \citep{heinz11,avcu17},  is more powerful than the attested phonological typology. Subjects in the artificial grammar learning paradigm  are, for example, able to successfully learn alternations that never surface in natural languages \citep{glewwe17,glewwe18,avcu18,begus18,begusDiss}. Second, overgeneration is a less severe violation than undergeneration. Absence of unnattested patterns that are derivable within a theory can be explained with external factors, such as historical developments or articulatory limitations. Not generating attested patterns, however, is a more serious shortcoming: a model of phonology should at minimum derive the observed phonological processes. Finally, the main reason the proposed model overgenerates is because the current proposal involves no information about the articulatory mechanism in speech production. In other words,  the GAN model is completely unconstrained for articulatory mechanisms. 

This would be problematic if the goal of the current model  were a network that models phonetic and phonological learning both on the articulatory and the cognitive levels. The aims of the current proposal, however, are more restricted.  The network models learning without any articulatory information. Lack of articulatory information in the model (and consequently, the overgeneration problem) might in fact be an advantage for computational models of the cognitive basis of speech production and perception.  It is likely that speech acquisition involves various different types of learning. Learning of motor-planning on the articulatory level is likely different from learning of articulatory targets based on perception, which is in turn likely controlled by other systems than learning of abstract symbol manipulation on the phonological level, even though these levels are interconnected in acquisition. Among the evidence that exemplifies the different levels of representation  are aphasia patients with different production errors  \citep{buchwald12}. If impairment targets the motor-planning unit, the phonological level is intact and the production error causes only deletion of [s] in \#sTV target clusters with the stop being unaspirated, as predicted by phonology. If, on the other hand, phonological computation is impaired, the stop surfaces as aspirated, similar to the outputs of our GAN model. By excluding articulatory information, we model phonetic and phonological learning as if they were unconstrained by articulators and therefore only influenced by the neural network architecture. In other words, we model phonological computation on a cognitive level as if no articulatory constraints were present in human speech. This is highly desired for the task of distinguishing those aspects of phonology that are influenced by cognitive factors from those that are influenced by articulation, motor planning, or historical developments \citep{begus18}. 

While the proposal in this paper does not directly address the discussion between generative  and exemplar-based approaches to phonology, the GAN models have the potential to offer some insights into this discussion as well. The results of the computational experiments suggest that the network learns to output data consistent with the training data without grammar-specific assumptions, which would support the exemplar-based approaches to phonology. On the other hand, the Generator network does seem to compress phonological information in its latent space in a way that does not directly correspond to stored exemplars. Further explorations of the latent space should shed light on this long-standing discussion.

Several further explorations and improvements of the model are warranted. The acoustic speech data fed to the network is modeled as waveform data points, i.e. pressure points in a time continuum (as proposed for WaveGAN in \citealt{donahue19}). This has considerable advantages for exploring the properties of the network, because spectral analysis introduces significant losses in the signal. A GAN trained on spectral transformations would likely be closer to reality, as human auditory mechanisms resemble spectral information more closely than raw pressure points \citep{young08,pasley12,mesgarani14}. Adding an articulatory model would likewise yield novel information on the role of articulatory learning on phonetic and phonological computation.

\section{Conclusion}
\label{conclusion}

The results of this paper suggest that we can model phonology not only with rules (as in rule-based approaches; \citealt{ch68}), exemplars \citep{p01}, finite-state automata \citep{heinz10,chandlee14}, input-output optimization (as in Optimality Theory; \citealt{ps93}), or with neural network architecture that already assumes some level of abstraction (see Section \ref{intro}), but as a mapping between random latent variables and output data in deep neural networks that are trained in an unsupervised manner from raw acoustic data. To the author's knowledge, this is the first paper testing learning of allophonic distributions in an unsupervised manner from raw acoustic data using neural networks and the first proposal to use GANs for modeling language acquisition. The Generative Adversarial model of phonology (trained on an implementation of DCGAN architecture for audio data in \citealt{donahue19}) derives outputs that resemble speech from latent variables.  The results of the computational experiment suggest that the network learns the conditional allophonic distribution of VOT duration. We propose a technique that identifies variables in the latent space that correspond to phonetic and phonological properties in the output, such as the presence of [s], and show that by manipulating these values, we can generate data with or without [s] in the output as well as control its  intensity and spectral properties of its frication noise.  While at least seven latent variables control the presence of [s], each of them likely has a phonetic function that controls spectral properties of the frication noise. The proposed technique thus suggests that the Generator network learns to encode phonetic and phonological information in its latent space. Finally, the model generates innovative outputs, suggesting its productive nature. The behavior of the model is compared against speech acquisition, speech errors, and speech impairment; several parallels are identified.

The current proposal models one allophonic distribution in English.  Training GAN networks on further processes and on languages other than English as well as probing the networks at different training steps should yield more information about learning representations of different features, phonetic and phonological processes, and about computational models of the cognitive aspects of human speech production and perception in general. This paper outlines a methodology for establishing internal representations and testing predictions against generated data, but represents  just a first step in  a broader task of modeling phonetic and phonological learning  in a Generative Adversarial framework.

The proposed model also has implications beyond modeling the cognitive basis of human speech. The results of establishing internal representations of the Generator network have implications for more applicable tasks in natural language processing. Identifying latent variables that correspond to output sounds allows for a model that generates desired input strings with different output properties. Discussing the details of such models is beyond the scope of this paper.

\section*{Conflict of Interest Statement}

The author declare that the research was conducted in the absence of any commercial or financial relationships that could be construed as a potential conflict of interest.

\section*{Author Contributions}

The author confirms being the sole contributor of this work and has approved it for publication.

\section*{Funding}
This research was funded by a grant to new faculty at the University of Washington.

\section*{Acknowledgments}
I would like to thank  Sameer Arshad for slicing data from the TIMIT database and Heather Morrison for annotating data. Parts of this research were published in \cite{begusscil}. All mistakes are my own.

\section*{Data Availability Statement}
The raw data supporting the conclusions of this manuscript will be made available by the author, without undue reservation, to any qualified researcher.

\bibliographystyle{frontiersinSCNS_ENG_HUMS} 
\bibliography{begusGANbib.bib}

\appendix

\section{Training data: Gamma regression}
\label{tdgr}

To confirm the presence of the durational distribution in VOT between the \#TV and \#sTV sequences in the training data, VOT durations based on TIMIT's manual annotations were measured across the two conditions. VOT in TIMIT is annotated from the release of the stop to the onset of the following vowel. Slices for which no VOT duration exists (only closure duration that includes the VOT) were excluded from this analysis, but were included in the training; altogether 47 sequences were thus excluded.  While the TIMIT database is occasionally misaligned, the errors are minor and likely do not crucially affect the outcomes. Table \ref{durations} and Figure \ref{violin1} summarize raw VOT durations across three places of articulation. Speaker identity is not included in the model, because it is irrelevant for the purpose of training a GAN network.

To test the significance of the presence of [s] as a predictor of VOT duration, the data were fit to a Gamma regression model (with log-link) with two predictors: \textsc{Structure} (the presence vs.~absence of [s]) and \textsc{Place} of articulation of the target stop (with three levels --- [p], [t], [k]) and their interaction. \textsc{Structure} was treatment-coded (with absence of [s] as the reference level), while \textsc{place} of articulation of the stop was sum-coded (with [k] as reference). The interaction term is significant (AIC = 46873.25 vs.~46885.73), which is why it is kept in the final model. The model (see Table \ref{timitlm}) shows that at the mean of the \textsc{Place} of articulation as a predictor, VOT is significantly shorter if T is preceded by [s] ($\beta=-0.84,  t =-49.69, p<0.0001$). Fitted values for \#TV are 56.97 ms [56.41,  57.53] ms  and for \#sTV 24.62 ms [23.86, 25.41]. The difference between the means is 32.35 ms.   The ratio of VOT durations (estimated with the \emph{emmeans} package; \citealt{emmeans}) between the two conditions ($\frac{\#TV}{\#sTV}$) equals $2.34$, ($\text{SE}=0.039$). Figure \ref{violin1} illustrates the significant difference and its magnitude between the two conditions across the three places of articulation. The significant interaction \#sTV:[t] is not informative  for our purposes.

\begin{table}
\centering
\begin{tabular}{lrrrr}
  \hline
 & Estimate & Std. Error & t value & Pr($>$$|$t$|$) \\ 
  \hline

(Intercept) &4.0426 & 0.0050 & 804.3778 & 0.0000 \\ 
  \#TV vs.~\#sTV & -0.8389 & 0.0169 & -49.6883 & 0.0000 \\ 
$[$p] vs.~mean  & -0.1383 & 0.0079 & -17.5160 & 0.0000 \\ 
  $[$t] vs.~mean & -0.0311 & 0.0068 & -4.5706 & 0.0000 \\ 
  \#sTV:[p] &  -0.1026 & 0.0257 & -3.9979 & 0.0001 \\ 
  \#sTV:[t]  &  0.0694 & 0.0214 & 3.2432 & 0.0012 \\ 
   \hline
\end{tabular}\caption{\label{timitlm}Coefficients of a Gamma regression model with duration of VOT in the training data as the dependent variable and condition (\#TV vs.~\#sTV) and \textsc{Place} of articulation (with interaction) as independent variables.}
\end{table}

\section{Generated data: Gamma regression}
\label{gammaregapp}

To test the significance of the observed distribution in the generated data, the data were fit to a Gamma regression model with VOT duration as the dependent variable and only one predictor: the presence of [s] (\textsc{Structure}). Place of articulation and following vowel were not added in the model, because they are often difficult to recover. \textsc{Structure} is a significant predictor of VOT duration:  $\beta=-0.46,  t=  -9.17, p<0.0001$. \footnote{Estimates for Intercept (when no [s] precedes) are $\beta=-2.79, t= -78.34, p<0.0001$.} Fitted values for \#TV are 61.47 ms with 95\% CI [57.33, 65.91] and for \#sTV 38.68 ms with 95\% CI [36.06, 41.50]. The difference between the means is 22.79 ms. The ratio of VOT durations (estimated with \emph{emmeans} package; \citealt{emmeans}) between the two conditions ($\frac{\#TV}{\#sTV}$) equals $1.59\ (\text{SE} = 0.080)$. 

\section{Latent variables:  Lasso regression and Random Forest models}
\label{lrrf}

To further test the accuracy of the regression model (in Section \ref{regression}) in identifying the variables that correlate with [s] in the output, the data were also fit to Lasso regression and Random Forest models. The same seven variables are also identified as having the highest estimates in a Lasso regression for binomial data, estimated with the \emph{glmnet} package \citep{glmnet} with cross-validated lambda values. Almost identical results are also derived with the Balanced Random Forest approach (estimated in \emph{randomForest} package in \citealt{randomforest}). The seven variables have the highest Mean decrease Gini estimates in a random forest model after 2,500 trees and with 9 variables randomly sampled for each tree. There is again a substantial decrease in estimates after the seven values. Mean decrease accuracy gives a similar ranking, with the exception that $z_5$ is the 8th highest predictor and $z_{74}$  the 18th highest. The accuracy of this estimate is highly variable with the choice of number of variables sampled and likely not as accurate as the regression models (possibly due to the fact that error rate for the presence of [s] group is high in the model --- 74.2\%). The value of variables were chosen based on smallest OOB error rate (tried on a range from 9 to 15 with 2,500 trees). We sample 271 variables from each group (the presence vs.~absence of [s]) each time to correct for the unbalanced sample.

\section{Generative test 2}
\label{supgen2}

The proportions from the Generative test 2 (Section \ref{gen2}) were fit to a beta regression linear model\footnote{Generalized additive models do not provide a better fit and in none of the six models is a smooth significantly different from a linear line.} (using \emph{mgcv} package; \citealt{mgcv}). The independent variables are estimates of the regression models in Figure \ref{gans} for each of the 31 variables tested. In fact, we can test which of the six regression models (from generalized additive to linear logistic regression) makes the best predictions about the latent variables and the correlation of the variables with the presence of  [s] in the output.  Six models were fit, one for each of the six regression models presented above (Full, Select, Modified, Excluded, Linear, LinearExcluded; Table \ref{gans}). The best-fitting model was chosen based on AIC: estimates of $z$-variables in the Linear model (Figure \ref{gans}) make the best predictions regarding the presence or absence of [s] in the output as tested with this independent generative approach. 

\section{Interpolation}
\label{supinter}
Number of knots for the model in Table \ref{zValueGam} is chosen as the default in the smooth term and as 5 in the random smooths. There is negative autocorrelation at lag 1, but with so little variance left unexplained (99.5\%; adjusted $R^2=0.99$), this likely does not affect outcomes substantially \citep{soskuthy17}. Autocorrelation is reduced when the ratio is modeled as normally distributed and correction for \textsc{ar(1)} correlation is added to the model with $\rho=0.98$ \citep{baayen16}. This, however, introduces a substantially worse fit. Since estimates of the smoothing terms are similar (with the same smooths being significant), we keep the beta regression model with autocorrelation.

\begin{table}[ht]
\centering
\begin{tabular}{lrrrr}
   \hline
     \multicolumn{5}{c}{\textbf{\textsc{Select}}}\\
A. parametric coef. & Estimate & Std. Error & t-value & p-value \\\hline
  (Intercept) & -5.3046 & 0.2104 & -25.2179 & $<$ 0.0001 \\ 
   \hline
B. smooth terms & edf & Ref.df & F-value & p-value \\ \hline
  s($z_{5}$) & 0.9828 & 9.0 & 57.0935 &  0.0000 \\ 
  s($z_{11}$) & 0.9823 & 9.0 & 55.4790 &  0.0000 \\ 
  s($z_{14}$) & 0.9791 & 9.0 & 46.7389 &  0.0000 \\ 
  s($z_{26}$) & 0.9802 & 9.0 & 49.5906 &  0.0000 \\ 
  s($z_{29}$) & 1.6222 & 9.0 & 51.2550 &  0.0000 \\ 
  s($z_{49}$) & 0.9819 & 9.0 & 54.1608 &  0.0000 \\ 
  s($z_{74}$) & 2.3630 & 9.0 & 50.3333 &  0.0000 \\ 
   \hline
  
  \hline
  
  \multicolumn{5}{c}{\textbf{\textsc{Linear excluded}}}\\
 & Estimate & Std. Error & z-value & Pr($>$$|$z$|$) \\ 
  \hline
(Intercept) & -6.1378 & 0.2879 & -21.32 & 0.0000 \\ 

  $z_{5}$ & 1.3678 & 0.1770 & 7.73 & 0.0000 \\ 
  $z_{11}$ & -1.3619 & 0.1725 & -7.89 & 0.0000 \\ 
   $z_{14}$ & 1.2739 & 0.1759 & 7.24 & 0.0000 \\ 
  $z_{26}$& 1.2932 & 0.1725 & 7.50 & 0.0000 \\ 
  $z_{29}$ & -1.3234 & 0.1705 & -7.76 & 0.0000 \\ 
  $z_{49}$ & -1.3557 & 0.1747 & -7.76 & 0.0000 \\ 
  $z_{74}$& -1.3280 & 0.1795 & -7.40 & 0.0000 \\ 
     \hline
\end{tabular}
\caption{Coefficients of the seven predictors with highest $\chi^2$ values or highest slope estimates from two models: \textsc{Select} and \textsc{Linear excluded}.} 
\label{tab.gam}
\end{table}

\begin{table}
\centering
\begin{tabular}{lrrrr}
      \hline
A. parametric coefficients & Estimate & Std. Error & t-value & p-value \\ \hline
  (Intercept) = $z_{11}$& -0.0571 & 0.0156 & -3.6505 & 0.0003 \\ 
      $z_5$ & -0.0404 & 0.0123 & -3.2820 & 0.0011 \\ 
  $z_{14}$& -0.0011 & 0.0144 & -0.0753 & 0.9400 \\ 
  $z_{26}$ & -0.0097 & 0.0115 & -0.8444 & 0.3989 \\ 
  $z_{29}$  & -0.0590 & 0.0113 & -5.2131 & $<$ 0.0001 \\ 
  $z_{49}$ & 0.0074 & 0.0112 & 0.6595 & 0.5100 \\ 

  $z_{74}$ & -0.0741 & 0.0121 & -6.1071 & $<$ 0.0001 \\ 
   \hline
B. smooth terms & edf & Ref.df & F-value & p-value \\ \hline
  s(zValuePerc):$z_5$ & 1.0002 & 1.0000 & 11.8417 & 0.0006 \\ 
  s(zValuePerc):$z_{11}$ & 4.1696 & 4.8546 & 14.1190 & $<$ 0.0001 \\ 
  s(zValuePerc):$z_{14}$& 5.3322 & 6.1117 & 36.6899 & $<$ 0.0001 \\ 
  s(zValuePerc):$z_{26}$ & 1.0003 & 1.0002 & 12.5952 & 0.0004 \\ 
  s(zValuePerc):$z_{29}$ & 1.0002 & 1.0000 & 12.0036 & 0.0006 \\ 
  s(zValuePerc):$z_{49}$ & 4.2002 & 4.8650 & 19.1225 & $<$ 0.0001 \\ 

  s(zValuePerc):$z_{74}$ & 3.2768 & 3.7863 & 1.2326 & 0.2479 \\ 
  fs(zValuePerc,sameValues,m=1,k=5) & 110.6863 & 143.0000 & 6.1542 & $<$ 0.0001 \\ 
  fs(zValuePerc,trajectoryZ,m=1,k=5) & 558.2060 & 728.0000 & 56.9670 & $<$ 0.0001 \\ 
   \hline
   \end{tabular}
\caption{Coefficients of a beta regression generalized additive model with ratio of maximum intensity ([s] vs.~vowel) as the dependent variable.} 
\label{zValueGam}
\end{table}

\begin{table}
\centering
\begin{tabular}{lrrrr}
   \hline
A. parametric coefficients & Estimate & Std. Error & t-value & p-value \\ \hline
  (Intercept) = $z_{11}$ & 4751.7378 & 84.7008 & 56.1002 & $<$ 0.0001 \\ 
    $z_{5}$ & 218.6576 & 116.9490 & 1.8697 & 0.0618 \\
  $z_{14}$ & -236.4061 & 134.6301 & -1.7560 & 0.0793 \\ 
  $z_{26}$ & 195.2722 & 108.9736 & 1.7919 & 0.0734 \\ 
  $z_{29}$ & 103.6866 & 107.8602 & 0.9613 & 0.3366 \\ 
  $z_{49}$ & 17.6464 & 106.4109 & 0.1658 & 0.8683 \\ 
 
  $z_{74}$ & 108.7466 & 113.8531 & 0.9551 & 0.3397 \\ 
   \hline
B. smooth terms & edf & Ref.df & F-value & p-value \\ \hline
  s(zValuePerc) = $z_{11}$& 7.5348 & 7.9933 & 12.1238 & $<$ 0.0001 \\ 
    s(zValuePerc):$z_{5}$ & 4.5539 & 5.7457 & 2.9261 & 0.0081 \\ 
  s(zValuePerc):$z_{14}$ & 7.3604 & 8.3734 & 5.7228 & $<$ 0.0001 \\ 
  s(zValuePerc):$z_{26}$ & 5.5900 & 6.8683 & 3.8049 & 0.0005 \\ 
  s(zValuePerc):$z_{29}$ & 5.8536 & 7.1301 & 2.9198 & 0.0045 \\ 
  s(zValuePerc):$z_{49}$ & 4.4714 & 5.6434 & 1.8590 & 0.0803 \\ 

  s(zValuePerc):$z_{74}$ & 4.2765 & 5.4186 & 2.8162 & 0.0136 \\ 
  fs(zValuePerc,sameValues,m=1,k=10) & 143.4989 & 288.0000 & 1.0560 & $<$ 0.0001 \\ 
  fs(zValuePerc,trajectoryZ,m=1,k=7) & 168.3558 & 1120.0000 & 0.2032 & $<$ 0.0001 \\ 
   \hline
\end{tabular}
\caption{Coefficients of a generalized additive model with center of gravity as the dependent variable with the marginal value of $z$-variables (\textsc{strong}). The model was fit with correction for autocorrelation with $\rho=0.7$.} 
\label{zValueCOGGamARsum}
\end{table}

\begin{table}
\centering
\begin{tabular}{lrrrr}
   \hline
A. parametric coefficients & Estimate & Std. Error & t-value & p-value \\ \hline
  (Intercept) = $z_{11}$& 1.0675 & 0.1045 & 10.2167 & $<$ 0.0001 \\ 
    $z_{5}$ & -0.4521 & 0.1420 & -3.1842 & 0.0015 \\ 
  $z_{14}$ & 0.3405 & 0.1693 & 2.0105 & 0.0446 \\ 
  $z_{26}$ & -0.4434 & 0.1323 & -3.3517 & 0.0008 \\ 
  $z_{29}$ & -0.6225 & 0.1339 & -4.6502 & $<$ 0.0001 \\ 
  $z_{49}$ & 0.0431 & 0.1332 & 0.3234 & 0.7464 \\ 

  $z_{74}$ & -0.5129 & 0.1383 & -3.7077 & 0.0002 \\ 
   \hline
B. smooth terms & edf & Ref.df & F-value & p-value \\ \hline
  s(zValuePerc) = $z_{11}$& 3.3590 & 4.0455 & 4.4859 & 0.0013 \\ 
    s(zValuePerc):$z_{5}$ & 1.0001 & 1.0001 & 1.8978 & 0.1686 \\ 
  s(zValuePerc):$z_{14}$ & 5.7086 & 6.9165 & 2.9066 & 0.0054 \\ 
  s(zValuePerc):$z_{26}$ & 1.0000 & 1.0000 & 2.3717 & 0.1238 \\ 
  s(zValuePerc):$z_{29}$ & 2.2995 & 2.8348 & 1.4855 & 0.2361 \\ 
  s(zValuePerc):$z_{49}$ & 5.3523 & 6.5335 & 2.5656 & 0.0106 \\ 

  s(zValuePerc):$z_{74}$ & 1.0000 & 1.0000 & 0.1912 & 0.6620 \\ 
  fs(zValuePerc,sameValues,m=1,k=10) & 69.5866 & 288.0000 & 0.4214 & $<$ 0.0001 \\ 
  fs(zValuePerc,trajectoryZ,m=1,k=7) & 174.7382 & 1120.0000 & 0.2422 & $<$ 0.0001 \\ 
   \hline
\end{tabular}
\caption{Coefficients of a generalized additive model with kurtosis as the dependent variable with the marginal value of $z$-variables (\textsc{strong}). The model was fit with correction for autocorrelation with $\rho=0.2$.} 
\label{zValueKURTGamARsum}
\end{table}

\begin{table}
\centering
\begin{tabular}{lrrrr}
   \hline
A. parametric coefficients & Estimate & Std. Error & t-value & p-value \\ \hline
  (Intercept) = $z_{11}$& 0.2726 & 0.0841 & 3.2434 & 0.0012 \\ 
    $z_{5}$ & -0.2686 & 0.1197 & -2.2448 & 0.0249 \\ 
  $z_{14}$ & -0.0188 & 0.1377 & -0.1368 & 0.8912 \\ 
  $z_{26}$ & -0.1965 & 0.1115 & -1.7629 & 0.0781 \\ 
  $z_{29}$ & -0.2011 & 0.1101 & -1.8270 & 0.0679 \\ 
  $z_{49}$ & -0.0403 & 0.1063 & -0.3792 & 0.7046 \\ 

  $z_{74}$ & -0.2468 & 0.1165 & -2.1193 & 0.0342 \\ 
   \hline
B. smooth terms & edf & Ref.df & F-value & p-value \\ \hline
  s(zValuePerc)  = $z_{11}$ & 4.5857 & 5.4215 & 1.3591 & 0.3433 \\ 
    s(zValuePerc):$z_{5}$ & 4.2885 & 5.5104 & 2.0917 & 0.0864 \\ 
  s(zValuePerc):$z_{14}$ & 6.4497 & 7.7372 & 3.3262 & 0.0009 \\ 
  s(zValuePerc):$z_{26}$ & 6.4653 & 7.7452 & 2.2045 & 0.0303 \\ 
  s(zValuePerc):$z_{29}$ & 3.8520 & 4.9849 & 2.0158 & 0.0716 \\ 
  s(zValuePerc):$z_{49}$ & 1.0000 & 1.0001 & 0.0105 & 0.9186 \\ 

  s(zValuePerc):$z_{74}$ & 4.0239 & 5.1943 & 1.9009 & 0.0916 \\ 
  fs(zValuePerc,sameValues,m=1,k=10) & 113.6068 & 288.0000 & 0.6943 & $<$ 0.0001 \\ 
  fs(zValuePerc,trajectoryZ,m=1,k=7) & 0.0001 & 1120.0000 & 0.0000 & 0.9908 \\ 
   \hline
\end{tabular}
\caption{Coefficients of a generalized additive model with skew as the dependent variable with the marginal value of $z$-variables (\textsc{strong}). The model was fit with correction for autocorrelation with $\rho=0.7$.} 
\label{zValueSKEWGamARsum}
\end{table}

\begin{table}
\centering
\begin{tabular}{lrrrr}
   \hline
A. parametric coefficients & Estimate & Std. Error & t-value & p-value \\ \hline
  (Intercept) = $z_{11}$& 4396.2895 & 88.8182 & 49.4976 & $<$ 0.0001 \\ 
    $z_5$ & 2.4059 & 85.3386 & 0.0282 & 0.9775 \\ 
  $z_{14}$ & 109.3881 & 101.5196 & 1.0775 & 0.2815 \\ 
  $z_{26}$ & 98.1943 & 79.3503 & 1.2375 & 0.2162 \\ 
  $z_{29}$ & -34.1064 & 78.4139 & -0.4350 & 0.6637 \\ 
  $z_{49}$ & -42.5635 & 77.1872 & -0.5514 & 0.5815 \\ 

  $z_{74}$ & 19.5268 & 85.7248 & 0.2278 & 0.8199 \\ 
   \hline
B. smooth terms & edf & Ref.df & F-value & p-value \\ \hline
  s(zValuePerc) = $z_{11}$ & 6.9763 & 7.4135 & 16.1815 & $<$ 0.0001 \\ 
    s(zValuePerc):$z_5$ & 1.0002 & 1.0003 & 0.0007 & 0.9793 \\ 
  s(zValuePerc):$z_{14}$ & 2.1327 & 2.4962 & 1.0245 & 0.2793 \\ 
  s(zValuePerc):$z_{26}$ & 1.0072 & 1.0110 & 0.2178 & 0.6468 \\ 
  s(zValuePerc):$z_{29}$ & 1.0003 & 1.0004 & 0.4048 & 0.5249 \\ 
  s(zValuePerc):$z_{49}$ & 1.0001 & 1.0002 & 3.5280 & 0.0606 \\ 

  s(zValuePerc):$z_{74}$ & 2.4946 & 2.9500 & 1.2008 & 0.2568 \\ 
  fs(zValuePerc,sameValues,m=1,k=10) & 198.6356 & 288.0000 & 3.3009 & $<$ 0.0001 \\ 
  fs(zValuePerc,trajectoryZ,m=1,k=7) & 413.0783 & 1120.0000 & 1.6191 & $<$ 0.0001 \\ 
   \hline
\end{tabular}
\caption{ Coefficients of a generalized additive model with center of gravity as the dependent variable with the value of $z$-variables at the point before [s] ceases from the output (\textsc{weak}).} 
\label{zValueKURTGamsum1}
\end{table}

\begin{table}
\centering
\begin{tabular}{lrrrr}
   \hline
A. parametric coefficients & Estimate & Std. Error & t-value & p-value \\ \hline
  (Intercept) = $z_{11}$& 0.6420 & 0.1544 & 4.1575 & $<$ 0.0001 \\ 
    $z_5$ & -0.0037 & 0.1463 & -0.0256 & 0.9796 \\ 
  $z_{14}$ & -0.4010 & 0.1685 & -2.3803 & 0.0174 \\ 
  $z_{26}$ & 0.0230 & 0.1368 & 0.1678 & 0.8668 \\ 
  $z_{29}$ & 0.0909 & 0.1344 & 0.6762 & 0.4991 \\ 
  $z_{49}$ & 0.0870 & 0.1323 & 0.6576 & 0.5109 \\ 

  $z_{74}$ & 0.1577 & 0.1429 & 1.1032 & 0.2702 \\ 
   \hline
B. smooth terms & edf & Ref.df & F-value & p-value \\ \hline
  s(zValuePerc) = $z_{11}$& 2.6481 & 2.9134 & 1.5639 & 0.2107 \\ 
    s(zValuePerc):$z_5$ & 1.0000 & 1.0000 & 3.2961 & 0.0697 \\ 
  s(zValuePerc):$z_{14}$ & 1.0000 & 1.0001 & 0.5569 & 0.4556 \\ 
  s(zValuePerc):$z_{26}$ & 1.9006 & 2.3489 & 1.0773 & 0.3078 \\ 
  s(zValuePerc):$z_{29}$ & 1.0000 & 1.0000 & 0.0284 & 0.8661 \\ 
  s(zValuePerc):$z_{49}$ & 1.0000 & 1.0000 & 0.0002 & 0.9887 \\ 

  s(zValuePerc):$z_{74}$ & 1.3675 & 1.6177 & 0.3165 & 0.5648 \\ 
  fs(zValuePerc,sameValues,m=1,k=10) & 181.3987 & 288.0000 & 2.3885 & $<$ 0.0001 \\ 
  fs(zValuePerc,trajectoryZ,m=1,k=7) & 128.9479 & 1120.0000 & 0.1673 & $<$ 0.0001 \\ 
   \hline
\end{tabular}
\caption{Coefficients of a generalized additive model with kurtosis as the dependent variable with the value of $z$-variables at the point before [s] ceases from the output (\textsc{weak}). The model was fit with correction for autocorrelation with $\rho=0.2$.} 
\label{zValueKURTGamARsum1}
\end{table}

\begin{table}
\centering
\begin{tabular}{lrrrr}
   \hline
A. parametric coefficients & Estimate & Std. Error & t-value & p-value \\ \hline
  (Intercept) = $z_{11}$& 0.2432 & 0.0734 & 3.3145 & 0.0009 \\ 
    $z_5$ & -0.0384 & 0.0758 & -0.5067 & 0.6125 \\ 
  $z_{14}$ & -0.0906 & 0.0873 & -1.0373 & 0.2998 \\ 
  $z_{26}$ & -0.1433 & 0.0705 & -2.0325 & 0.0423 \\ 
  $z_{29}$ & -0.0392 & 0.0698 & -0.5613 & 0.5747 \\ 
  $z_{49}$ & -0.0191 & 0.0687 & -0.2777 & 0.7813 \\ 

  $z_{74}$ & -0.0151 & 0.0740 & -0.2043 & 0.8381 \\ 
   \hline
B. smooth terms & edf & Ref.df & F-value & p-value \\ \hline
  s(zValuePerc) = $z_{11}$& 5.2698 & 5.9125 & 3.3712 & 0.0037 \\ 
    s(zValuePerc):$z_5$ & 1.0000 & 1.0000 & 0.5871 & 0.4437 \\ 
  s(zValuePerc):$z_{14}$ & 1.0000 & 1.0000 & 1.7508 & 0.1860 \\ 
  s(zValuePerc):$z_{26}$ & 1.0000 & 1.0000 & 0.1276 & 0.7210 \\ 
  s(zValuePerc):$z_{29}$ & 1.5340 & 1.8995 & 0.3881 & 0.6718 \\ 
  s(zValuePerc):$z_{49}$ & 1.0000 & 1.0000 & 0.5952 & 0.4406 \\ 

  s(zValuePerc):$z_{74}$ & 2.1616 & 2.7315 & 0.7639 & 0.3898 \\ 
  fs(zValuePerc,sameValues,m=1,k=10) & 170.1493 & 288.0000 & 2.2288 & $<$ 0.0001 \\ 
  fs(zValuePerc,trajectoryZ,m=1,k=7) & 47.7523 & 1120.0000 & 0.0661 & 0.0001 \\ 
   \hline
\end{tabular}
\caption{Coefficients of a generalized additive model with skew as the dependent variable with the value of $z$-variables at the point before [s] ceases from the output (\textsc{weak}). The model was fit with correction for autocorrelation with $\rho=0.3$.} 
\label{zValueSKEWGamARsum1}
\end{table}

 \begin{figure}
\centering
\includegraphics[width=0.8\textwidth]{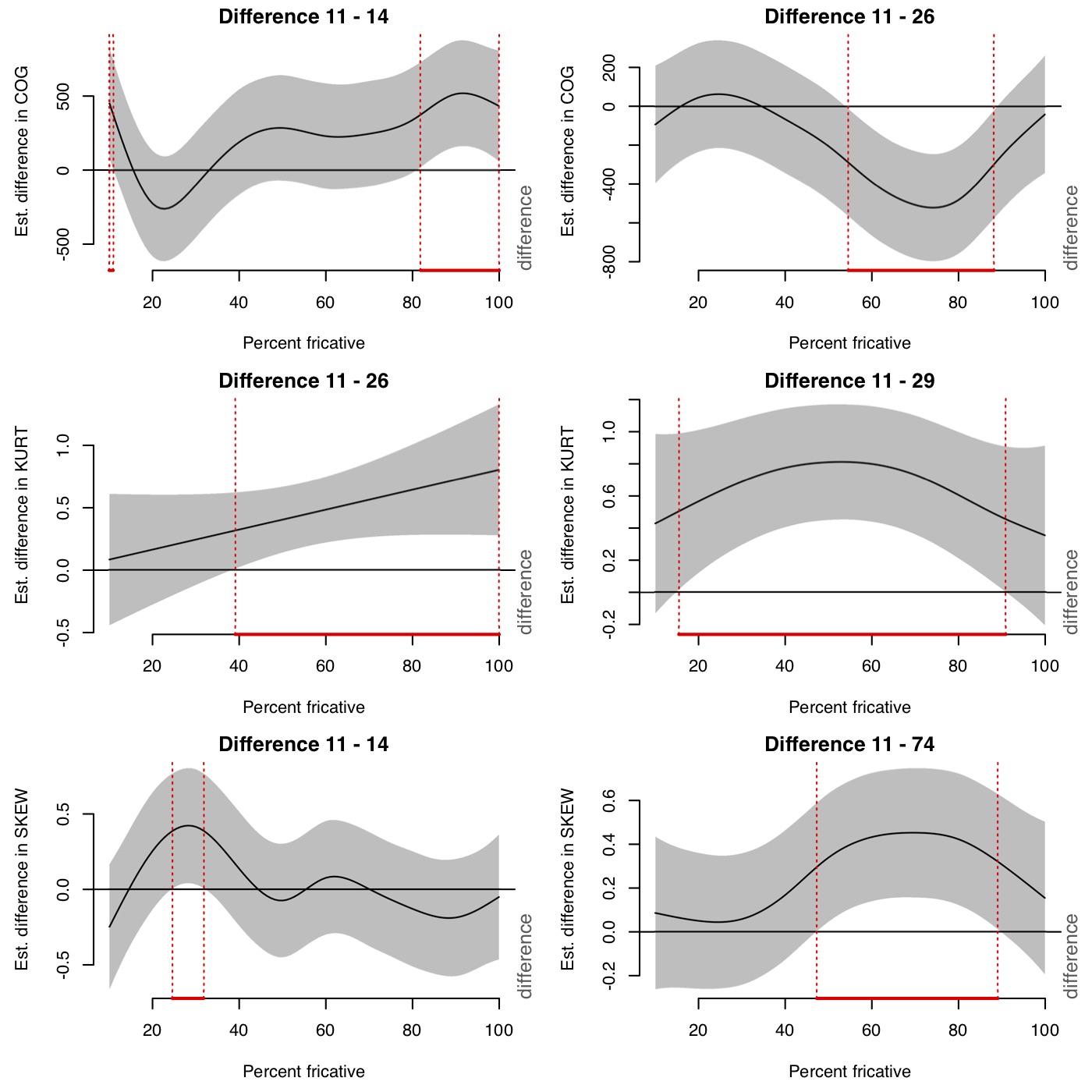}
\caption{\label{differences}Pairwise difference smooths in COG, kurtosis, and skew   between $z_{11}$ and other two variables for models in Figure \ref{kurt}.}
\end{figure}



\end{document}